\documentclass[pdflatex,sn-nature,iicol]{sn-jnl}% Nature Portfolio (npj Computational Materials), two-column journal layout

\usepackage{graphicx}%
\usepackage{amsmath,amssymb,amsfonts}%
\usepackage{mathtools}%
\usepackage{xcolor}%
\usepackage{textcomp}%
\usepackage{manyfoot}%
\usepackage{booktabs}%
\usepackage{microtype}%
\usepackage[title]{appendix}%

\makeatletter
\renewcommand{\@@corrauthor}[2][]{\def\@authfrstarg{#1}\@corauemailtrue%
  \advance\corraucount by 1%
  \g@addto@macro\artauthors{%
    \global\@auemailtrue%
    \Authorfont%
    \def\baselinestretch{1}%
    \authorsep{#2}\unskip\textsuperscript{*}\unskip%
    \def\authorsep{\au@and~}%
    \global\let\sep\@empty\global\let\@corref\@empty%
  }}%
\makeatother

\begin{document}

\title[Transfer Learning Architectures for Scalable Multi-Fidelity Bayesian
Optimization]{Transfer Learning Architectures for Scalable Multi-Fidelity Bayesian
Optimization}

\author{\fnm{Jaewook} \sur{Lee}}

\author{\fnm{Ethan} \sur{Errington}}

\author{\fnm{Christian D.} \sur{Lorenz}}

\author*{\fnm{Miao} \sur{Guo}}\email{miao.guo@kcl.ac.uk}

\affil{\orgdiv{Department of Engineering}, \orgname{King's College London},
\orgaddress{\city{London}, \postcode{WC2R 2LS}, \country{United Kingdom}}}
\abstract{\unboldmath Self-driving laboratories increasingly rely on multi-fidelity Bayesian optimization (MFBO) to balance cheap, approximate evaluations against scarce, expensive ones, with a predictive surrogate at its core. Gaussian processes (GPs) are the default choice, but they scale poorly as data accumulate and assume a smooth landscape that molecular and materials search spaces routinely violate. Transfer learning offers an alternative suited to this regime: it learns a representation from abundant cheap data and adapts it to sparse expensive data. Despite its use in property prediction, transfer learning has not been tested as the engine of a closed-loop optimization. Here we benchmark eleven transfer-learning surrogates against four GP methods under an identical selection rule, fidelity budget, and model size, across nine tasks spanning synthetic functions to real chemistry and materials problems. GPs win on smooth, low-dimensional functions but perform worst on molecular and materials problems, where transfer-learning surrogates reach substantially better solutions using far less computation. Because acquisition policy is held fixed across surrogates, this advantage is attributable to the surrogate itself. Uncertainty-driven exploration is not reliably beneficial, and calibration does not predict optimization performance, so greedy exploitation of the transfer-learned mean is the more robust default. Transfer learning is therefore the surrogate of choice for molecular and materials MFBO.}

\keywords{Multi-fidelity Bayesian optimization, Transfer learning,
Materials and molecular discovery}

\maketitle

\section{Introduction}

Self-driving laboratories (SDLs) are transforming scientific discovery by integrating automated experimentation with AI-driven decision-making in closed-loop optimization workflows. In SDL systems, experiments are designed, executed and analysed autonomously, while the derived data iteratively update a surrogate model that guides subsequent experimentation\citep{abolhasani2023rise,
hickman2025atlas}. As the decision-making core of the SDL, the surrogate model and acquisition policy largely determine the efficiency of discovery. In chemistry and
materials science, where experimental search spaces are vast and evaluations are expensive, Bayesian
optimization (BO) has therefore become the dominant approach for identifying optimal conditions using minimal experimental efforts\citep{shahriari2015taking}.

A central limitation of conventional BO is its reliance on a single source of information, which imposes a trade-off between evaluation cost and prediction accuracy. Scientific and engineering problems naturally possess hierarchical sources of information spanning multiple fidelities \citep{peherstorfer2018survey}. Low-fidelity (LF) evaluations, including approximate simulations, reduced-order process models, coarse-grained molecular dynamics or simplified computational fluid dynamics, are computationally inexpensive but often noisy, biased or approximate. High-fidelity (HF) experiments and rigorous simulations provide substantially greater accuracy, but at far higher cost and lower throughput. Multi-fidelity Bayesian optimization (MFBO) addresses this challenge by combining these complementary information sources within a unified optimization framework, enabling  inexpensive LF evaluations to broadly explore the
design space while reserving scarce HF measurements for validation of the most promising
candidates~\citep{kennedy2000predicting}. MFBO has therefore emerged as an increasingly powerful strategy for
improving sample efficiency in autonomous discovery under constrained experimental budgets~\citep{sabanza2025best}.

\begin{figure*}[tp]
\centering
%% Display item 1 of 6 — composite: (a) closed-loop overview; (b--k) final
%% regret on the nine benchmarks; (l--n) family-split anytime-AUC advantage row
%% (standalone export of the grid's middle row; the complete grid with
%% final-regret and marginal rows is Supplementary Fig.~\ref{suppfig:family_split}).
%% Panel letters are baked into the PDFs (unique across the composite); the
%% overview PNG carries no letter, so "a" is overlaid in LaTeX below.
\makebox[0.9\textwidth][l]{\fontsize{8}{9}\selectfont\textsf{\textbf{a}}}\\[1pt]
\includegraphics[width=0.9\textwidth]{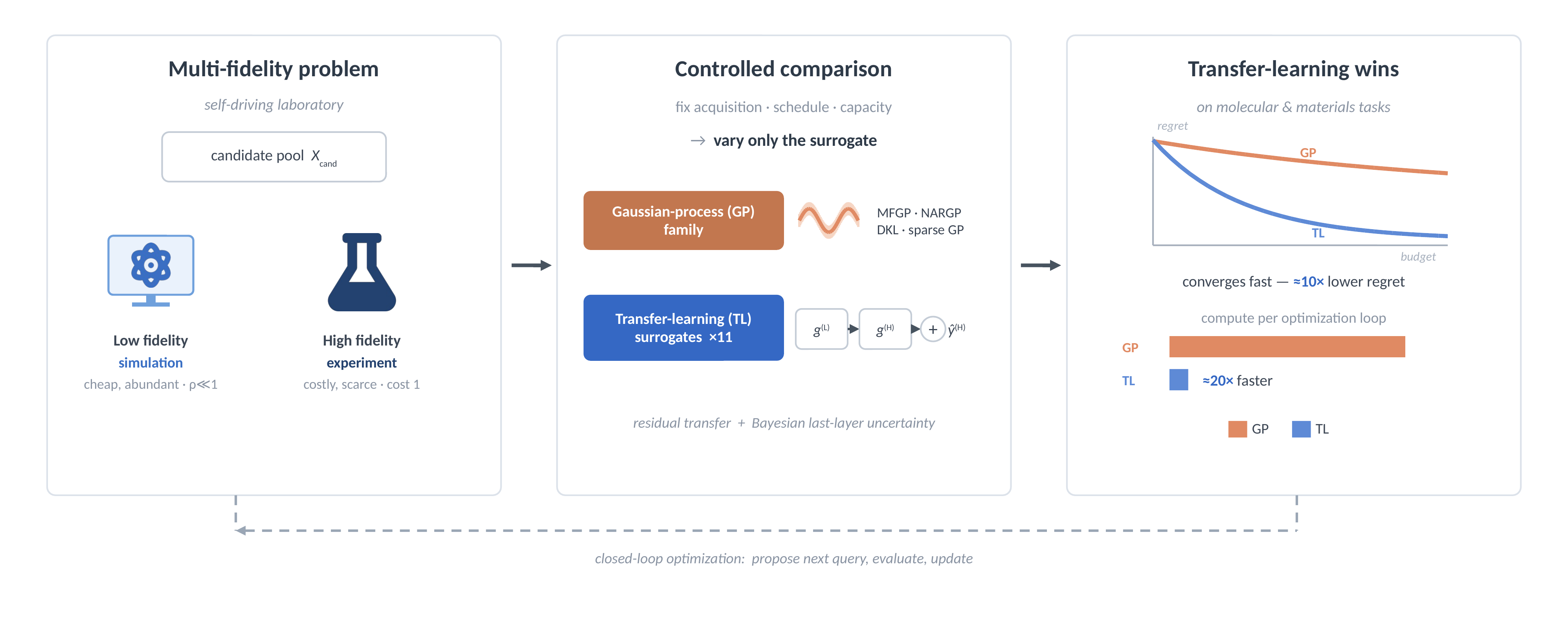}\\[4pt]
\includegraphics[width=\textwidth]{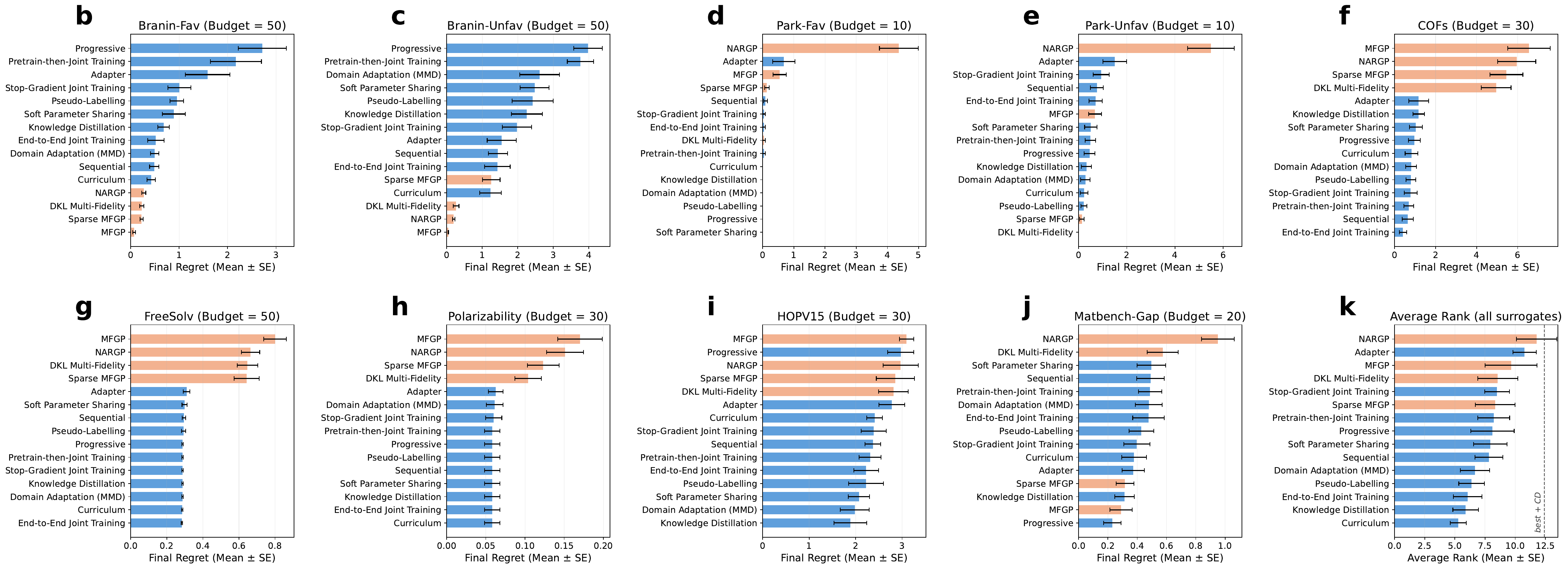}\\[4pt]
\includegraphics[width=\textwidth]{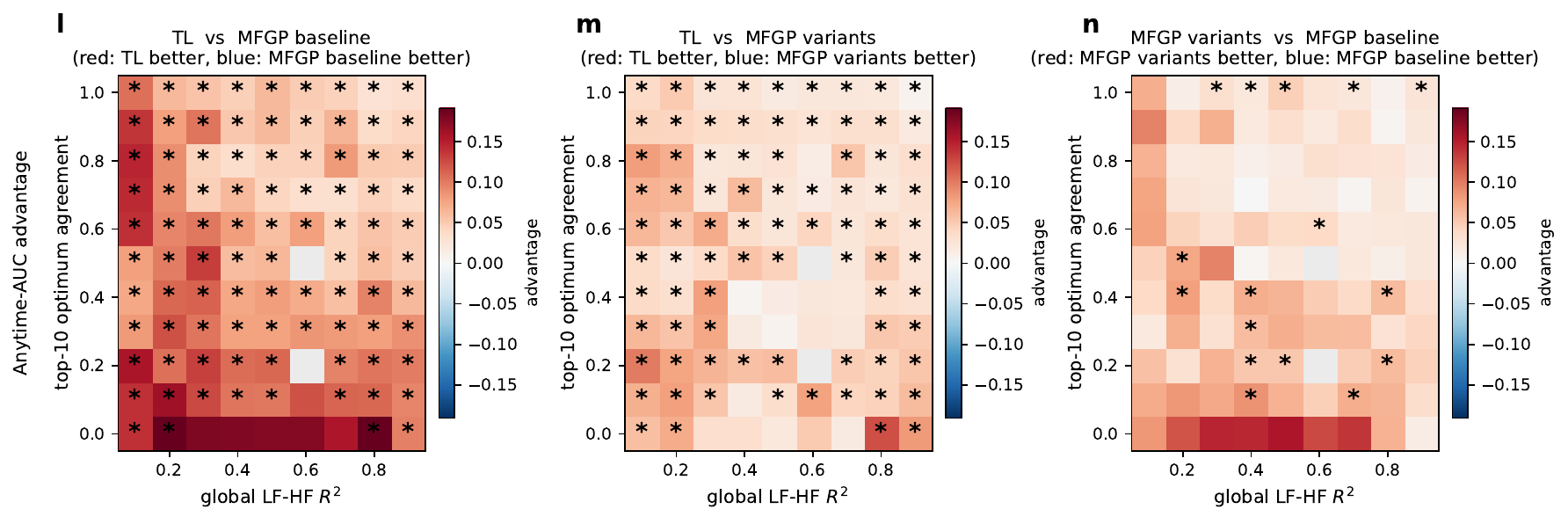}
\caption{Overview of the surrogate comparison and headline results.
\textbf{a,} in a self-driving laboratory, a shared candidate pool
$\mathcal{X}_{\mathrm{cand}}$ is queried through an abundant, cheap low-fidelity (LF)
simulation (cost $\rho \ll 1$) and a scarce, expensive high-fidelity (HF) experiment
(cost $1$). With the same selection rule (acquisition policy), the same mix of cheap and expensive
evaluations, and all transfer-learning (TL) surrogates the same size, only the surrogate is
varied: a four-member Gaussian-process family (the baseline MFGP, NARGP, a deep-kernel
GP, and a sparse variational MFGP) against eleven TL surrogates, in which a network trained on
the cheap LF data, $g^{(L)}$, is corrected by a second network, $g^{(H)}$, trained on
the accurate HF data.
\textbf{b--k,} final regret at the designated budget (mean $\pm$ s.e.\ over 20
seeds), surrogates sorted within each panel (lower is better); Gaussian-process family
in salmon, TL surrogates in blue. The Gaussian-process family attains the lowest regret
on the smooth, low-dimensional Branin functions (\textbf{b},\textbf{c}) but the highest
on the molecular-descriptor benchmarks (\textbf{f}--\textbf{i}); on the
low-discrimination Matbench-gap task (\textbf{j}) the baseline MFGP is competitive.
Panel \textbf{k} reports the average rank of all fifteen surrogates across the nine
benchmarks; the dashed line marks the best average rank plus the Nemenyi critical
difference ($\alpha = 0.05$)~\citep{demsar2006statistical}, which no surrogate pair
exceeds, so the ordering is descriptive (omnibus rank analysis in Supplementary
Note~\ref{suppnote:rank_stats}).
\textbf{l--n,} family-split anytime-advantage maps from the controlled
fidelity-quality grid (Methods). Mean anytime advantage (area under the regret--budget
curve) in bins of the global LF--HF $R^2$ (horizontal) and the top-10 optimum agreement
between fidelities (vertical), comparing TL against the baseline MFGP (\textbf{l}), TL
against the MFGP variants (\textbf{m}), and the MFGP variants against the baseline MFGP
(\textbf{n}); red favours the first-named family and asterisks mark bins significant
under a per-bin stratified signed-rank test with Benjamini--Hochberg
false-discovery-rate correction across each map (Methods). The TL
advantage over the baseline is positive in every cell, grows as the top-10 agreement
falls, and is nearly flat in $R^2$; the complete grid, with the final-regret maps and
marginal profiles, is Supplementary Fig.~\ref{suppfig:family_split}.}
\label{fig:overview}\label{fig:final_regret}\label{fig:family_heatmap}
\end{figure*}

Most MFBO methods remain rooted in Gaussian process (GP) surrogates whose strength lies in calibrated uncertainty estimation. Standard multi-fidelity Gaussian processes (MFGP) perform strongly in low-data regimes and provide uncertainty estimates, but their limitations become increasingly apparent
 in the high-dimensional and heterogeneous search spaces characteristic of molecular and materials discovery.  GP surrogates scale poorly as LF data accumulate~\citep{williams2006gaussian}, while kernel-based similarity assumptions can struggle to capture nonlinear, biased or location-dependent relationships between LF and HF signals~\citep{snoek2015scalable, seeger2004gaussian, perdikaris2017nonlinear}. This limitation is particularly important in SDLs, where LF sources are often imperfect experimental proxies: they may be strongly informative only in restricted regions of the design space, misleading near the optimum or weakly correlated with the final HF objective.

Here we test an alternative hypothesis: in high-dimensional multi-fidelity discovery, optimization performance is governed less by uncertainty calibration or global LF--HF correlation than by the surrogate's capability to learn useful representations of the design landscape. When LF and HF signals share structure, such representations support cross-fidelity transfer; when LF information is weak or biased, they can still improve optimization by providing a more expressive model of high-dimensional molecular and materials descriptors than fixed-kernel GPs. This reframes multi-fidelity optimization as a representation-learning problem, in which the central challenge is to identify which structures in inexpensive data remain informative for expensive HF decisions.

Transfer-learning surrogates provide a natural mechanism for this reframing. By learning representations
from abundant LF data and adapting them to sparse HF observations through targeted correction~\citep{yosinski2014transferable}, these models can capture cross-fidelity structure while retaining the expressive capacity needed for \emph{prediction} in complex molecular and materials descriptor spaces~\citep{buterez2024transfer}. This flexibility is particularly important when fixed-kernel GP surrogates are poorly matched to nonlinear, heterogeneous or locally biased LF--HF relationships. Although transfer-learning models have advanced multi-fidelity molecular property prediction, their role as decision-making engines in closed-loop optimization remains largely unexplored~\citep{li2020dnnmfbo}. In molecular and materials discovery, MFBO has largely been developed around GP-based
surrogates~\citep{sabanza2025best}, whereas transfer-learning architectures representing distinct 
mechanisms of knowledge transfer have not yet been systematically evaluated under controlled closed-loop optimization conditions. It therefore remains unclear when transfer-learning surrogates outperform GP baselines, which transfer mechanisms are most effective and whether any advantage arises from better uncertainty estimates, better predictive means or improved representation of the design space.

We address these questions through a controlled benchmark study of 15 multi-fidelity surrogates across nine synthetic, chemistry and materials optimization tasks. The comparison includes eleven transfer-learning
surrogates and four GP baselines: autoregressive
MFGP~\citep{kennedy2000predicting}, nonlinear autoregressive GP (NARGP)~\citep{perdikaris2017nonlinear},
deep-kernel GP~\citep{wilson2016deep}, and sparse variational MFGP\citep{titsias2009variational, hensman2013gaussian}. All transfer-learning surrogates shared a two-stage structure in which  an LF network learns from abundant proxy data and an HF network learns the residual correction required to predict sparse HF observations\citep{gawlikowski2023survey}.
The architectures differ only in how information is transferred between these stages, allowing the transfer mechanism to be isolated from model capacity. We evaluated all methods under a shared closed-loop optimization protocol and used matched-acquisition control to separate surrogate effects from acquisition-policy effects (Fig.~\ref{fig:overview}).

Our results show that GPs remain highly effective when their inductive bias matches the objective, but that transfer-learning surrogates are more effective on most molecular and materials benchmarks considered here. The advantage is explained primarily by the quality of the transfer-learned predictive mean and by preservation of the near-optimal region, rather than by calibrated uncertainty or global LF--HF correlation alone. Transfer-learning surrogates also provide more favourable computational scaling under repeated model updates, a regime typical of SDL workflows with abundant LF data. These findings suggest that surrogate design for MFBO should place greater emphasis on scalable learned representations that can transfer information across fidelities while adapting to biased, local or weak LF signals.

%The significance of this reframing reaches beyond a single model comparison. Treating the
%model as something that learns and carries a representation from cheap LF data to accurate HF data
%changes where effort in a self-driving laboratory is best spent, toward the model and the
%quality of that representation rather than toward ever more elaborate rules for choosing
%experiments, and because it concerns the model alone it can be adopted in existing automated
%platforms without changing how they choose what to run. The same idea applies well beyond our
%benchmarks: any discovery problem that pairs a cheap, approximate LF test with a scarce,
%expensive HF measurement, as catalysis, drug discovery, materials screening, and process design
%routinely do, fits the same template. And as learned representations of molecules and
%materials keep improving, a model built to learn and transfer a representation is well placed
%to benefit from them where a fixed, hand-chosen model is not, making this approach a durable
%foundation for the decision-making engine at the heart of autonomous discovery.

\section{Results}\label{sec:results}

We evaluated fifteen surrogates on nine benchmarks
(Table~\ref{tab:benchmarks}) under a shared fixed protocol (Methods). Performance was measured as simple regret on HF
evaluations as a function of cumulative cost, averaged over twenty random seeds.

\subsection{Gaussian processes perform best on smooth, low-dimensional
objectives}\label{sec:res-branin}

The baseline MFGP (the standard autoregressive multi-fidelity GP; Methods) achieved the lowest final regret among all fifteen surrogates on both
Branin scenarios, indicating that GP models remain highly effective when
their inductive bias matches the objective structure. On Branin-Fav and Branin-Unfav (the favourable and unfavourable fidelity scenarios; Methods) the baseline MFGP reached
 mean final regrets of $0.07$ and $0.05$, respectively, ranking first in
both cases, whereas the best transfer-learning surrogate reached final regrets of $0.43$ and $1.24$
(Fig.~\ref{fig:final_regret}; paired Wilcoxon $p = 1.0\times10^{-3}$ and
$1.3\times10^{-4}$, matched-pairs rank-biserial $r = -0.79$ and $-1.00$, where $r > 0$
favours transfer learning; Methods). This observation is consistent with the Mat\'{e}rn-5/2 kernel providing a strong smoothness prior for an analytically smooth, two-dimensional objective,
allowing the GP posterior to model the surface accurately from relatively few observations.
The advantage was specific to the Branin family and did not extend across all synthetic benchmarks. On the four-dimensional Park functions, simple regret rapidly approached zero for most methods after only a small number of HF evaluations (Fig.~\ref{fig:final_regret}d,e), making these benchmarks weakly
 discriminative  at the budget considered. Several transfer-learning surrogates had already reached zero regret on Park-Fav whereas the baseline MFGP had not, placing it
in the lower half of the ranking on both Park scenarios. Therefore, the GP advantage on
synthetic objectives was confined to the smooth, two-dimensional Branin
functions, rather than synthetic benchmarks in general.
These results suggest that GPs are highly competitive when
the kernel prior is well matched to the structure of the objective, but that this advantage is problem-dependent rather than universal.

\subsection{Transfer-learning surrogates outperform on molecular and materials
benchmarks}\label{sec:res-chem}

On the molecular-descriptor benchmarks the ranking changed markedly. Whereas the baseline MFGP performed the best on the Branin functions it ranked at the bottom on covalent organic frameworks (COFs), solvation free energy (FreeSolv),
polarizability, and organic photovoltaics (HOPV15). The three MFGP variants (NARGP, the deep-kernel GP, and the sparse
variational MFGP) showed similar
high-regret behaviour on these benchmarks (Fig.~\ref{fig:final_regret}, salmon bars). On COFs the best-performing surrogate,
End-to-End Joint Training, reached a final regret of $0.41$ compared with $6.55$ for the baseline MFGP
($r = 0.97$, $p = 2.1\times10^{-4}$) and $4.9$--$6.0$ for the MFGP variants. The
corresponding values were $0.28$ versus $0.80$ on FreeSolv ($r = 1.00$,
$p = 1.9\times10^{-6}$), $0.06$ versus $0.17$ on polarizability ($r = 1.00$,
$p = 6.3\times10^{-4}$) and $1.89$ versus $3.09$ on HOPV15 ($r = 0.61$, $p = 0.020$);
effect sizes for all nine benchmarks are given in Supplementary
Table~\ref{supptab:effect_sizes}. Thus substituting the baseline MFGP with
nonlinear cross-fidelity coupling, jointly learned kernels, or  sparse
variational approximation did not remove the performance gap, suggesting that the limitation lies in the
GP modelling assumptions in these compressed descriptor spaces, rather than in a  single baseline implementation.

\begin{figure*}[t]
\centering
\includegraphics[width=\textwidth]{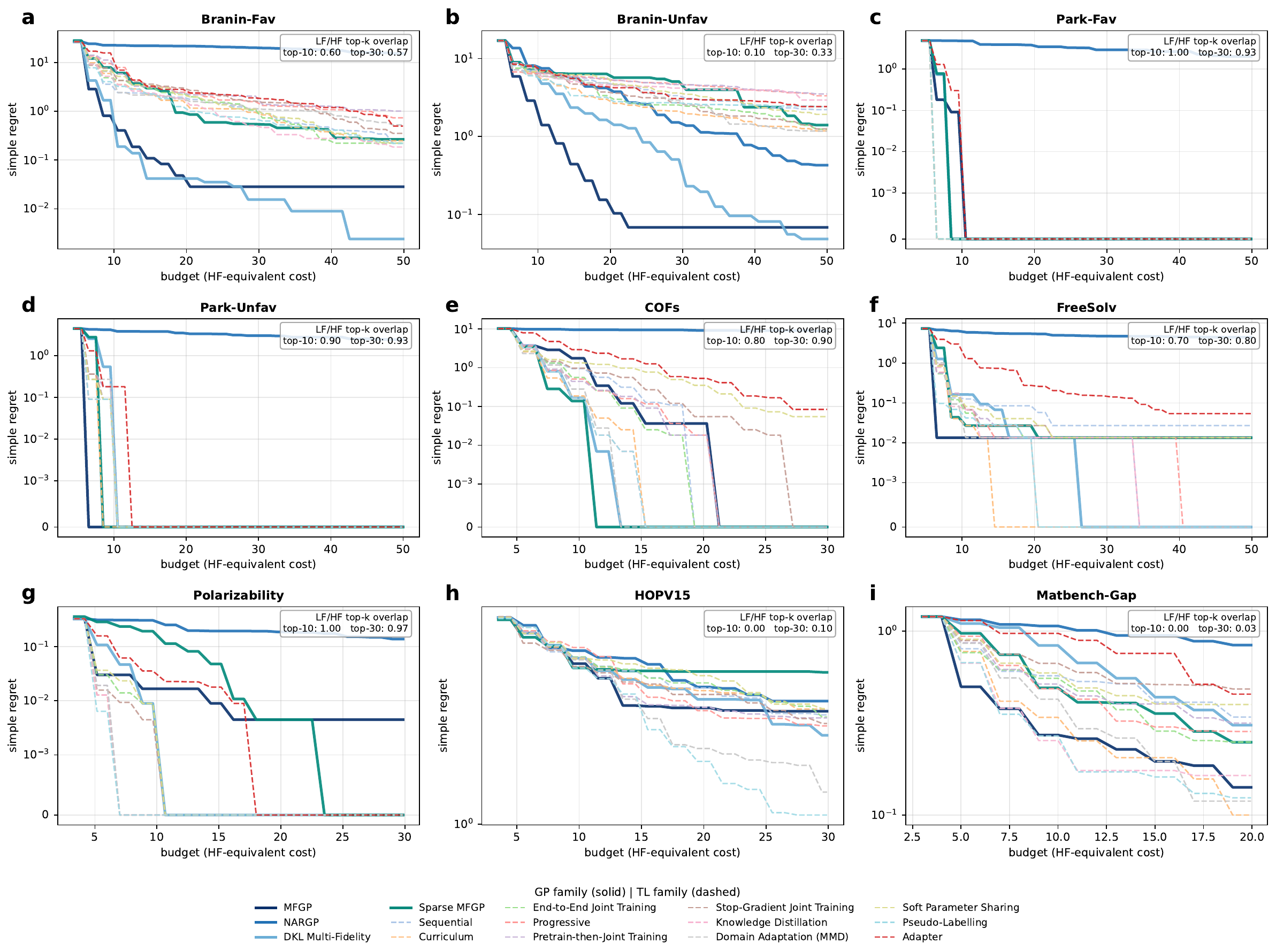}
\caption{Regret trajectories as a function of cumulative budget (mean $\pm$ s.e.\ over 20
seeds); Gaussian-process family in solid lines, transfer-learning surrogates in dashed
lines, one panel per benchmark. On the Branin family (\textbf{a},\textbf{b}) the
baseline MFGP leads throughout; on the chemistry benchmarks COFs, FreeSolv, and
polarizability (\textbf{e}--\textbf{g}) the Gaussian-process family remains in a
high-regret regime while transfer-learning surrogates continue to improve as the
budget grows. Insets report the low-/high-fidelity (LF/HF) top-$k$ overlap of each benchmark.}
\label{fig:regret_trajectory}
\end{figure*}

Two mechanisms explain this reversal. First, the Mat\'{e}rn-5/2 kernel imposes a
single global notion of smoothness, whereas the molecular descriptor space, compressed by principal
component analysis (PCA), can be non-stationary and the LF--HF discrepancy may vary strongly across input space, particularly near the optimum. In this setting, a stationary-kernel GP can misrank promising candidates, while a transfer-learning surrogate can learn task-adapted representations from the larger
LF dataset and model a smoother residual between fidelities. Second, the performance gap
was already visible at the first HF evaluation rather than emerging only after additional HF data were collected
(Fig.~\ref{fig:regret_trajectory}). This suggests that the bottleneck is representational (kernel
misspecification) rather than simply a consequence of HF data scarcity.  The conventional pattern where GPs perform best early and transfer-learning
surrogates improve only later was therefore not observed. The HOPV15 result further clarifies the source of the advantage. Because the LF signal is almost uninformative for this task
($R^2 = 0.02$), the improvement cannot be attributed to successful LF-to-HF transfer. Instead, it indicates that the transfer-learning surrogate provides
a better function approximator for the molecular descriptors than a stationary GP. Across the empirical benchmarks, the advantage of transfer-learning surrogates therefore has two complementary sources: improved representation learning, which remains beneficial even when the LF proxy is weak, and
transfer from abundant LF data, which becomes useful when the LF signal is informative, as in
COFs, FreeSolv, and polarizability.

Performance also depended on transfer mechanism. Curriculum, Knowledge Distillation
and End-to-End Joint Training achieved the lowest average ranks across the nine
benchmarks among all fifteen surrogates (Fig.~\ref{fig:final_regret}k). Because the
surrogate ranking reverses between the synthetic and molecular regimes, an omnibus
test on the pooled nine-benchmark ranking is not expected to resolve individual pairs,
and it does not (Supplementary Note~\ref{suppnote:rank_stats}); the per-benchmark
paired comparisons of Supplementary Table~\ref{supptab:effect_sizes}, rather than the
pooled ordering, carry the statistical support. The
spread of average ranks (from $5.3$ to $11.8$) nevertheless suggests that outcomes
depended not merely on using transfer learning but on how transfer was implemented.
In this benchmark suite, curriculum transfer, knowledge distillation and joint
training are the most dependable default choices.

Matbench-gap provides an important boundary case. In this inorganic
band gap benchmark, optimized towards a photovoltaic target of $1.4$~eV, the baseline MFGP was competitive, ranking second among all fifteen surrogates with a final regret of $0.29$ versus  $0.23$ for the best transfer-learning
surrogate ($r = 0.19$, $p = 0.47$; Fig.~\ref{fig:final_regret}j). This does not contradict the
advantage of the learned surrogates on harder molecular-descriptor benchmarks; rather, it reflects the limited discrimination of this task. The full regret range
across the baseline MFGP and the eleven transfer-learning surrogates was only $0.27$
compared with $6.1$ on COFs, and the baseline MFGP lay within $0.8$ standard errors of the best method. Thus,
surrogate choice had little practical effect on this benchmark. Two features explain the weak discrimination on Matbench-gap.
First, it had the smallest evaluation budget in the suite with only ten HF queries, a regime where the cubic scaling of GP inference is negligible and the GP prior can be especially influential. Second,  a near-optimal set was comparatively dense: $3.6\%$ of the $3{,}347$-compound
pool lay within $0.2$~eV of the target, whereas only $2$ of $608$ COFs lay within $1\%$ of
the optimum. Surrogate choice was therefore most consequential on the hard,
sharply-peaked molecular-descriptor benchmarks and less important on dense, small-budget tasks,
consistent with the difficulty analysis below.

\subsection{Optimum agreement, not global correlation, governs the transfer-learning
advantage}\label{sec:res-grid}

On the empirical benchmarks the global informativeness of the LF source and
its reliability near the optimum vary together, so either property could explain when
transfer learning wins. To separate them, we constructed a controlled grid on the
polarizability substrate in which curated LF sources vary the global LF--HF
$R^2$ and the top-10 optimum agreement, the overlap between the LF and HF
top-10 candidate sets, independently, giving $126$ conditions of ten seeds each under
the standard optimization protocol; the GP family was split into the
baseline MFGP and its variants, and the three families were compared pairwise on final
regret and on anytime performance, the area under the regret--budget curve (Methods).
The anytime-advantage maps are shown in Fig.~\ref{fig:family_heatmap}l--n and the
complete grid in Supplementary Fig.~\ref{suppfig:family_split}; per-cell effect-size
distributions are summarized in Supplementary Table~\ref{supptab:grid_effect_sizes}.

Three observations emerge. First, transfer learning dominates the baseline MFGP across
the entire plane: its anytime advantage is positive in $126$ of $126$ cells, with
$120$ of $126$ ($95\%$) individually significant after Benjamini--Hochberg
false-discovery-rate correction across the map (map-level Wilcoxon across cells,
$p = 2.0\times10^{-22}$; median per-cell rank-biserial $r = 1.00$, interquartile range
$1.00$--$1.00$, that is, in the typical cell every seed favours transfer learning;
Methods), and no cell reverses on final regret ($81$ of $126$ cells favour transfer
learning and the remaining $45$ tie at zero regret). Second,
the GP representative matters: against the best MFGP variant the
final-regret gap closes, with a median advantage near zero, but the anytime advantage
persists and remains individually significant in $101$ of $126$ cells ($80\%$) under
the same correction (median per-cell $r = 1.00$,
interquartile range $0.91$--$1.00$). Third, the advantage is organized
along the agreement axis rather than the correlation axis: at fixed $R^2$ it grows
steeply as the top-10 agreement falls, whereas at fixed agreement it is nearly flat
across the $R^2$ range (marginal profiles in Supplementary
Fig.~\ref{suppfig:family_split}). The property that decides when a transfer-learned
surrogate pays off is therefore not the global correlation commonly used to
characterize an LF source, but whether that source preserves the ordering of
the best candidates.

\subsection{The advantage stems from the surrogate, not the acquisition
policy}\label{sec:res-acquisition}

\begin{figure*}[t]
\centering
\includegraphics[width=0.92\textwidth]{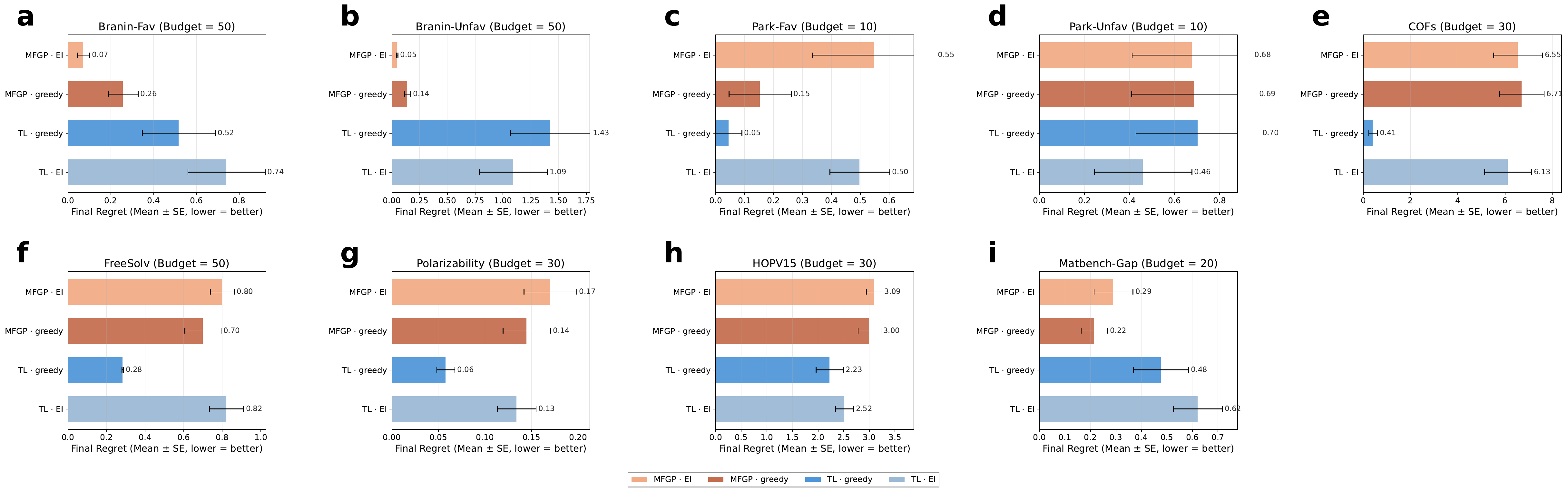}
\caption{Surrogate $\times$ acquisition comparison. Regret (mean $\pm$ s.e.) at the budget
shown in each panel title, for the multi-fidelity Gaussian-process (MFGP) and transfer-learning
(TL) surrogates under uncertainty-driven (expected improvement, EI) and greedy acquisition policies. The two Park functions
(\textbf{c},\textbf{d}) are solved by every condition at the full budget, so following
Fig.~\ref{fig:final_regret} they are evaluated at the earlier budget $B=10$, before every
condition saturates. On the chemistry benchmarks the two baseline MFGP bars are nearly equal,
showing that the acquisition policy does not explain the baseline MFGP's poor performance there,
whereas the transfer-learning surrogate degrades sharply when uncertainty-driven exploration replaces
greedy exploitation of the transfer-learned mean.}
\label{fig:acquisition}
\end{figure*}

Because the GP and transfer-learning surrogates used structurally different acquisition
policies under the default protocol, we matched the acquisition policy across both model families to test whether
the observed advantage of transfer-learning surrogates could be explained by this asymmetry. In the default setting, the baseline MFGP used an uncertainty-driven expected-improvement (EI) acquisition whereas the transfer-learning surrogates used a greedy policy (Methods). We therefore compared matched acquisition settings by running the baseline MFGP under a greedy policy
and the transfer-learning surrogate under an uncertainty-driven policy using its own Bayesian linear-regression (BLR) predictive
standard deviation (Fig.~\ref{fig:acquisition}). Using the greedy policy for the baseline MFGP had little effect on performance. Final regret changed only marginally, for example $6.60 \rightarrow 6.71$ on COFs and $0.13 \rightarrow
0.15$ on polarizability.  Thus matching the acquisition policy did not improve the baseline MFGP, indicating that its limitation arises primarily from the surrogate rather than the acquisition policy. Under this matched greedy policy the transfer-learning surrogate retained
a large advantage on COFs, achieving approximately an $8\times$ lower regret than the baseline MFGP and a $16\times$ lower regret for the best transfer-learning
surrogate. Smaller but consistent advantages were also observed on the remaining molecular benchmarks. The transfer-learning
surrogate showed the opposite behaviour: its performance was strongly sensitive to its acquisition policy. On COFs regret
increased from $0.88$ under the greedy policy to $6.13$ when its predictive uncertainty was used to drive
exploration.
These results suggest that the baseline MFGP
posterior \emph{mean} misranks the promising regions in these descriptor spaces, so changing the acquisition policy has limited effect. By contrast, the transfer-learning surrogate appeared to provide a more useful ranking of HF candidates, making performance more dependent on how scarce HF
evaluations were allocated between exploitation and exploration. We therefore examined this sensitivity using a broader acquisition-function portfolio.

\subsection{Uncertainty-driven exploration has benchmark-dependent value}\label{sec:res-uq}

\begin{figure*}[tp]
\centering
%% Display item 4 of 6 — composite: (a--d) acquisition portfolio;
%% (e--m) calibration versus regret. Both blocks belong to this section's
%% uncertainty story; panel letters are baked into the PDFs and unique across
%% the composite.
\includegraphics[width=0.88\textwidth]{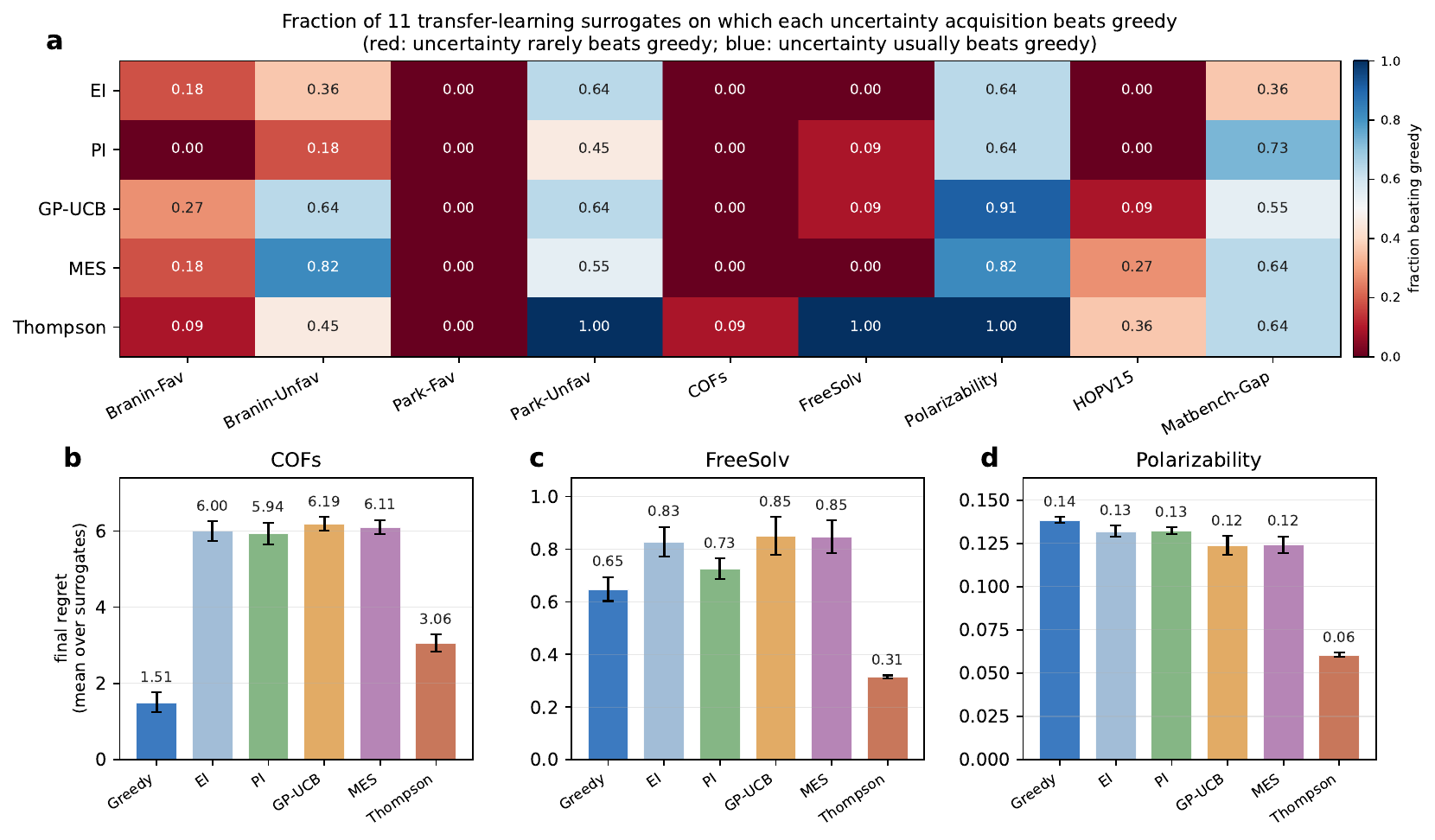}\\[6pt]
\includegraphics[width=\textwidth]{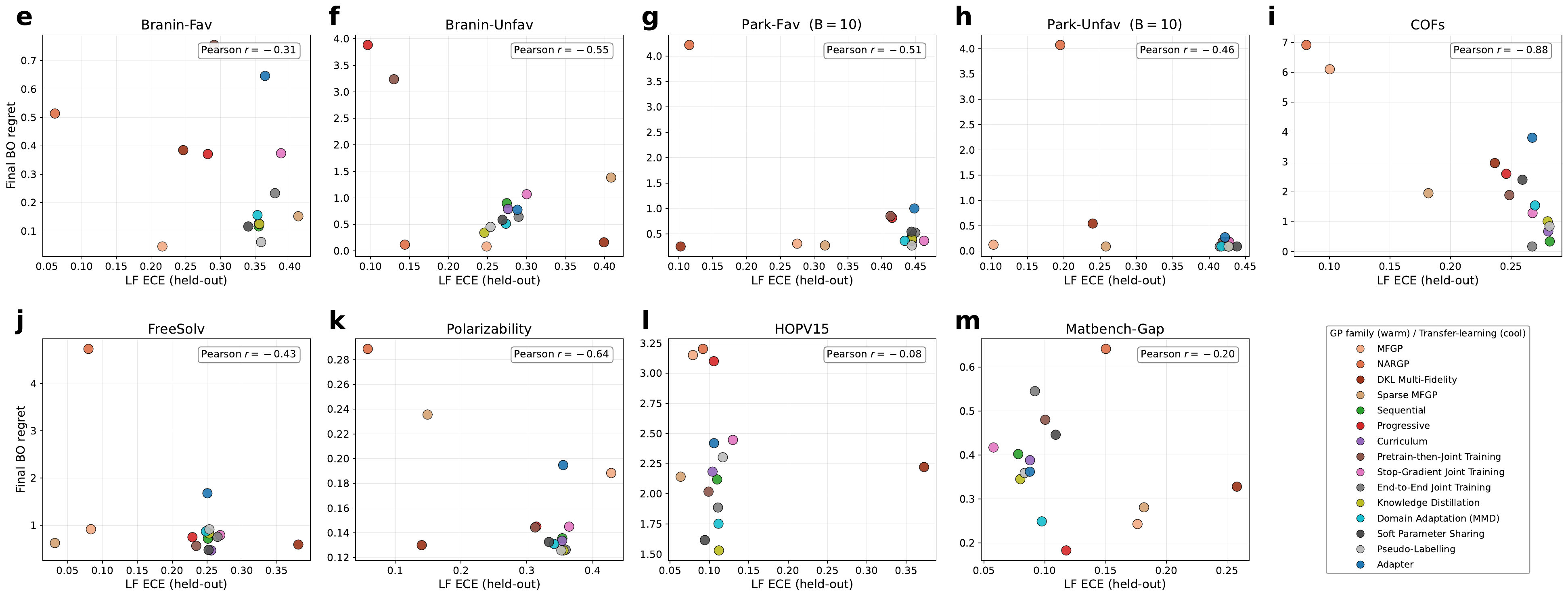}
\caption{The value of uncertainty-driven exploration is case-dependent, and calibration
does not predict optimization performance.
\textbf{a--d,} a five-member acquisition portfolio spanning the standard taxonomy
(expected improvement and probability of improvement; the Gaussian-process lower
confidence bound; max-value entropy search; Thompson sampling), each using the transfer-learning
surrogate's Bayesian linear-regression (BLR) predictive standard deviation, compared against the greedy policy
across all eleven transfer-learning surrogates and nine benchmarks. \textbf{a,} the
fraction of the eleven surrogates on which each uncertainty-driven acquisition beats the greedy policy (red:
rarely; blue: usually); averaged over benchmarks, Thompson sampling beats the greedy policy most
often and the improvement-based acquisitions least often. \textbf{b--d,} final regret (mean
over the eleven surrogates; error bars, s.e.\ across surrogates) on the three
high-correlation chemistry benchmarks (LF--HF rank correlation $\ge 0.94$): every
uncertainty-driven acquisition is worse than the greedy policy on COFs, whereas Thompson sampling beats
it on FreeSolv and polarizability.
\textbf{e--m,} calibration versus regret: each point is one of the fifteen surrogates (Gaussian-process
family in warm colours, transfer-learning surrogates in cool colours; legend); axes are
the low-fidelity expected calibration error and the final regret, one panel per
benchmark. The Pearson correlation is negative on every benchmark (from $-0.88$ on COFs
to $-0.08$ on HOPV15), so the best-calibrated surrogates (the Gaussian-process family)
are, if anything, the worst optimizers on the molecular-descriptor benchmarks. The two
Park benchmarks (\textbf{g},\textbf{h}) saturate to $\approx 0$ regret by the
full budget (Park-Fav additionally exhausts its $256$-point candidate pool), so both
are evaluated at the early decision budget $B=10$, consistent with the regret figures
(Figs.~\ref{fig:final_regret} and~\ref{fig:regret_trajectory}).}
\label{fig:portfolio}\label{fig:calibration}
\end{figure*}

Whether calibrated surrogate uncertainty improves BO depends on both the acquisition policy and the structure of the benchmark. To test this systematically, we
replaced the greedy policy with a representative portfolio of five
uncertainty-driven acquisitions spanning the standard acquisition-function taxonomy~\citep{shahriari2015taking}: EI~\citep{jones1998efficient} and
probability of improvement~\citep{kushner1964new} which are improvement-based; the
GP lower confidence bound~\citep{srinivas2012information} which is optimistic;
max-value entropy search~\citep{wang2017max} which is information-theoretic; and Thompson
sampling~\citep{russo2018tutorial} which is sampling-based. Each acquisition used the BLR
predictive standard deviation from the transfer-learning surrogate, while keeping the surrogate architecture, fidelity schedule, HF evaluation protocol and random seeds
fixed. The portfolio was evaluated across all eleven transfer-learning surrogates and nine
benchmarks (Fig.~\ref{fig:portfolio}).

The effect of uncertainty-driven exploration was case-dependent rather than uniform. On COFs, where the LF
signal transferred strongly,
 the greedy policy outperformed all five uncertainty-driven acquisitions for ten of the eleven
surrogates. Even the most robust uncertainty-driven acquisition approximately doubled the regret relative to the greedy policy, while the
improvement-based acquisitions increased regret severalfold. This suggests that in strong transfer settings, exploration could waste scarce
HF budget on uncertain but less promising candidates when the transfer-learned
mean already provides a useful ranking. The opposite pattern was observed on FreeSolv and polarizability. On both high-correlation chemistry
benchmarks, Thompson sampling outperformed the greedy policy for all eleven transfer-learning
surrogates, and on polarizability most 
uncertainty-driven acquisitions improved over the greedy policy. The acquisition family mattered more than the presence
of uncertainty alone. Thompson sampling was the most robust uncertainty-driven acquisition,
outperforming the greedy policy on the largest fraction of surrogates on average, whereas the
improvement-based acquisitions (EI and probability of improvement) were the least reliable (Fig.~\ref{fig:portfolio}a).
The greedy policy therefore remained the most robust single default,
whereas uncertainty-driven exploration showed benchmark-dependent benefits, motivating evaluation across an acquisition-function
portfolio rather than reliance on a single acquisition.

Calibration quality was a poor predictor of optimization performance: across all fifteen surrogates, the correlation between expected calibration error and final regret was negative on every benchmark, ranging from $-0.88$ on COFs to $-0.08$ on HOPV15 (Fig.~\ref{fig:calibration}e--m). If anything, the best-calibrated surrogates, including the entire GP family, were the weakest optimizers on the molecular-descriptor benchmarks, showing that better-calibrated uncertainty did not translate into better optimization. This is consistent with uncertainty acting as a secondary, benchmark-dependent factor rather than a primary driver of performance.
The bounded $\tanh$ activation used throughout was likewise justified by
optimization rather than calibration. In a
matched $\tanh$-versus-ReLU ablation on the molecular-descriptor benchmarks, calibration error was nearly identical for the two activations ($0.30$
versus
$0.30$), whereas  $\tanh$ achieved lower regret ($0.39$ versus $0.58$). Across the full benchmark suite,
the activation effect was benchmark-dependent and small relative to the surrogate effect
(Supplementary Information). Together, these results indicate that the advantage of transfer-learning surrogates arises primarily from the quality of the transfer-learned mean, with uncertainty-driven
exploration acting as a secondary and benchmark-dependent factor. 

\subsection{When is a surrogate necessary?}\label{sec:res-topk}

\begin{figure*}[tp]
\centering
%% Display item 5 of 6 — composite: (a,b) top-k overlap and LF-only screening
%% regret; (c) compute scaling law. Together: when a learned surrogate is
%% needed, and what it costs. Panel letters baked into the PDFs.
\includegraphics[width=0.92\textwidth]{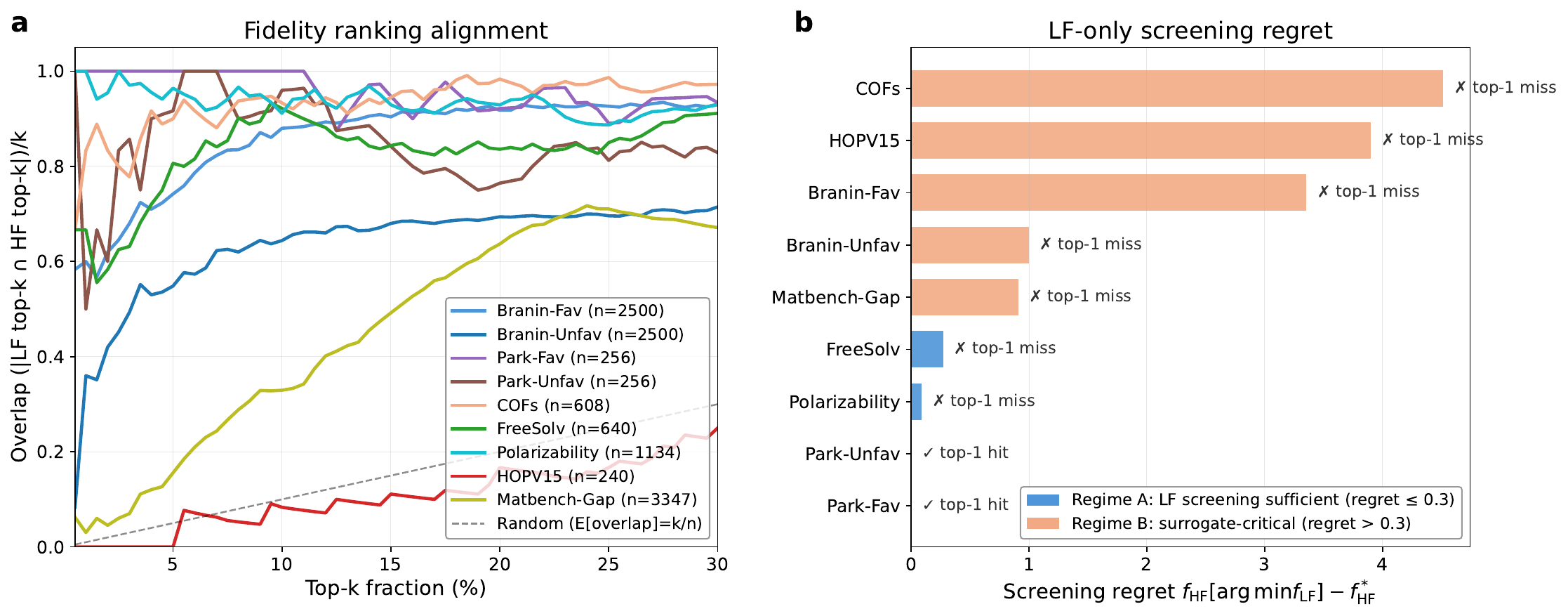}\\[6pt]
\includegraphics[width=0.85\textwidth]{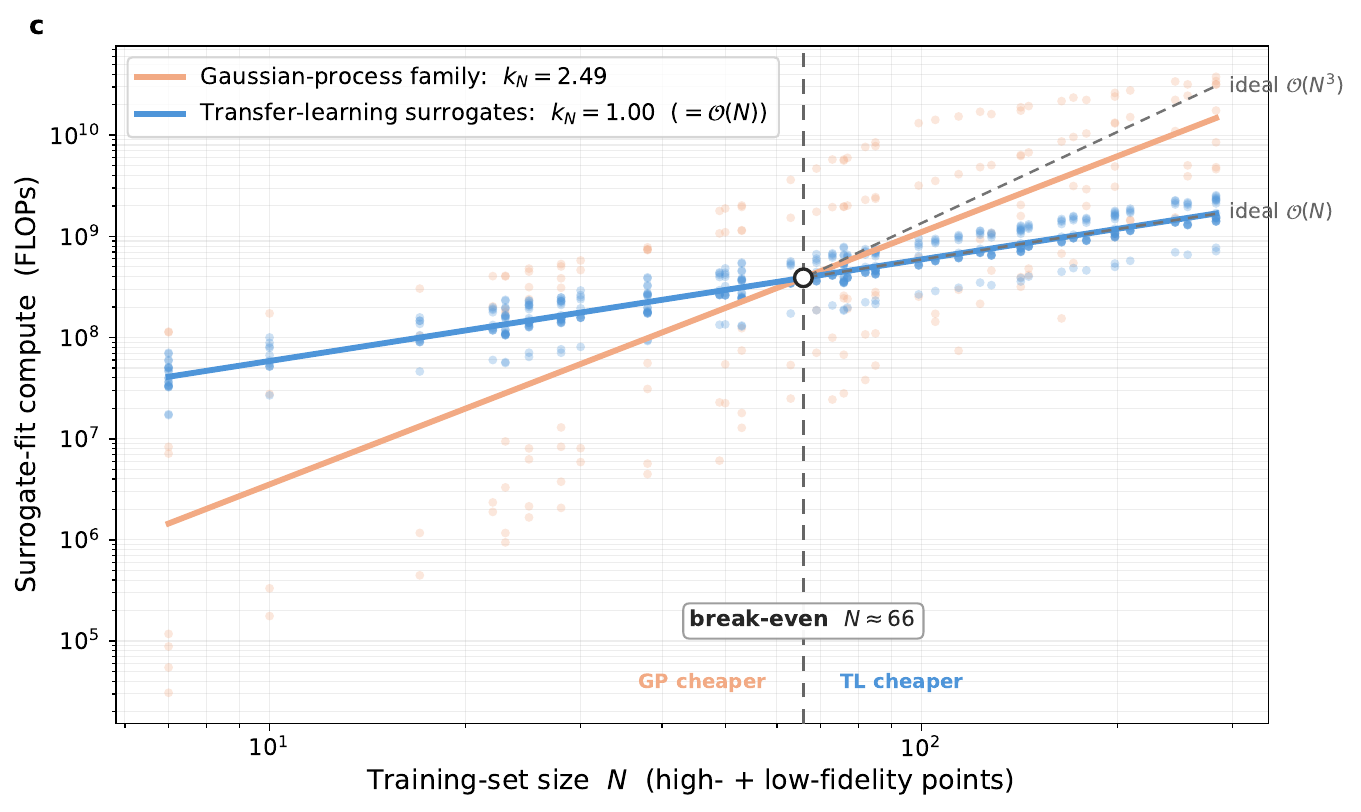}
\caption{When a surrogate is needed, and what it costs.
\textbf{a},\textbf{b,} task difficulty and the need for a surrogate. Top-$k$ overlap
between the low- and high-fidelity rankings (\textbf{a}) and the resulting LF-only
screening regret (\textbf{b}) for each benchmark. High global correlation does not
guarantee top-$k$ agreement, so a learned surrogate is required precisely where a
low-fidelity screen fails (Regime~B), while in Regime~A the screen is already
near-optimal.
\textbf{c,} computational scaling law across the nine benchmarks. Each point is
the median surrogate-fit compute (hardware-independent floating-point operations, FLOPs) of one surrogate on one
benchmark at one budget fraction, placed at the training-set size $N$ (high- plus
low-fidelity points) it was fitted on; salmon, Gaussian-process family; blue,
transfer-learning surrogates. A shared-slope power-law fit gives $k_N = 1.00$
($R^2 = 1.00$, exactly $\mathcal{O}(N)$) for the eleven transfer-learning surrogates
and $k_N = 2.49$ ($R^2 = 0.94$) for the Gaussian-process family, which lies below pure
cubic because it includes the scalable sparse variational MFGP, with the exact GPs (the
baseline MFGP and NARGP) approaching the ideal $\mathcal{O}(N^3)$ (faint dashed guides). The two family
laws cross at a break-even size $N \approx 66$ (vertical line): below it the Gaussian
processes are cheaper to fit, above it the transfer-learning surrogates are cheaper by
a margin that grows with $N$, reaching roughly $5\times$ at $N = 200$ and $10\times$ at
$N = 300$. Absolute per-benchmark FLOPs and the average compute ranking are in
Supplementary Fig.~\ref{suppfig:computing_time} (Supplementary
Note~\ref{suppnote:computational_cost}).}
\label{fig:topk}\label{fig:scaling_law}
\end{figure*}

A learned surrogate is not equally valuable for every optimization problem. To identify when surrogate learning adds benefit, we compared each method with a surrogate-free screening baseline that ranks candidates by their LF
value and evaluates the top-ranked candidates at HF. The resulting deterministic screening
regret provides a reference for distinguishing cases where the LF proxy is sufficient from those in which a learned cross-fidelity model is required
(Fig.~\ref{fig:topk}). This analysis separated the benchmarks into two regimes. In Regime~A, represented by Park-Fav and polarizability, the top LF
candidates coincide almost exactly with the top HF candidates. The top-1\% overlap was approximately $1$ and the screening regret was no greater than $ 0.09$. Under this setting, a learned surrogate added little value and all
methods, including non-learning baselines, performed similarly. In Regime~B, represented by COFs and
Branin, high global LF--HF correlation conceals poor agreement among the top-ranked  candidates. COFs had a
rank correlation of $0.998$ but a screening regret of $4.52$. In this setting, a learned surrogate was
essential; the best transfer-learning surrogate achieved a regret of $0.41$, substantially outperforming the surrogate-free screen. Regime membership could not be inferred from the global correlation alone. COFs and
polarizability both had $R^2 \approx 0.99$ but their screening regrets differed by approximately
$50\times$. Thus, the need for a surrogate depends less on global LF--HF correlation than on whether the LF proxy preserves the near-optimal region. The
transfer-learning surrogate was a robust choice as it matched the surrogate-free screen when the proxy already identified the optimum and substantially improved performance when the proxy failed near the top of the ranking. Both HOPV15 and Matbench-gap fall in
Regime~B. An LF screen produced a regret of $3.90$ on HOPV15 where the Scharber proxy
was nearly uninformative and $0.91$ on Matbench-gap. These results indicate that  surrogate learning is warranted,
even though the choice among surrogates had limited impact on the
low-discrimination Matbench-gap task.

\subsection{Computational cost}\label{sec:res-cost}

We quantified the computational cost of each surrogate in floating-point operations (FLOPs). FLOPs were counted for one surrogate fit and one pool-wide prediction and accumulated over the
optimization loop. This measure provides a hardware-independent
comparison of the scaling behaviours of the two model families. We report FLOPs rather than wall-clock time because it isolates algorithmic scaling from implementation and hardware effects; it does not capture memory footprint, which is particularly relevant for the sparse variational MFGP's inducing-point overhead, or engineering factors such as parallelization and library-level optimization that affect wall-clock cost in deployment.  The
eleven transfer-learning surrogates exhibited linear scaling with the number of observations with a measured
fit-FLOP exponent of $k = 1.0$ for cost versus training-set size. This reflects the use of fixed-architecture
neural network training. By contrast, the exact GPs, the baseline MFGP and NARGP, showed cubic scaling
($k = 3.0$), consistent with exact GP inference and marginal-likelihood optimization. 
The sparse variational MFGP and the deep-kernel GP occupied an intermediate computational regime (Fig.~\ref{fig:scaling_law}). The difference in scaling produced a clear crossover in computational cost. The fitted cost curves intersected at $N \approx 66$; below this point, the
GPs were cheaper to fit whereas above it  the transfer-learning surrogates became more efficient. Because the fitted scaling exponents differed by about $1.5$ between the model families, the computational gap widened with increasing data, 
reaching approximately $5$ times at $N = 200$ and $10$ times at $N = 300$.
Integrated over the optimization loop, the sparse variational MFGP
and the deep-kernel GP were the most expensive methods on every benchmark.
The exact GPs, the baseline MFGP and NARGP, became increasingly expensive as the evaluation
budget grew (Supplementary Note~\ref{suppnote:computational_cost}, Supplementary
Fig.~\ref{suppfig:computing_time}). The sparse variational MFGP required up to
approximately $45$-fold more FLOPs than the fastest transfer-learning surrogate, for example $5.0$ versus $0.11$ TFLOPs on FreeSolv. 
 The corresponding maximum gaps were approximately $25$-fold for the deep-kernel GP and $11$-fold for the baseline MFGP. On the smallest pools and budgets such as HOPV15 and Park-Unfav, the cubic cost of exact GPs remained negligible; the separation between model families increased as the number of observations grew. Therefore, under the fixed optimization protocol, 
 transfer-learning surrogates provided both lower regret on most molecular and materials benchmarks and more favourable computational scaling. This advantage is particularly relevant to SDL settings, where
abundant LF data and repeated surrogate updates can make model retraining cost an important component of the closed-loop workflows.

\section{Discussion}\label{sec:discussion}

By bringing transfer learning into the multi-fidelity optimization loop, we find a
transfer-learning surrogate to be the better engine for molecular and materials
discovery: within our protocol the choice of surrogate, more than the acquisition policy or
the fidelity schedule, governs MFBO performance in
high-dimensional molecular and materials spaces.
Across nine benchmarks, the baseline MFGP, like the broader GP family, was
either the best surrogate (smooth, low-dimensional Branin) or the worst (the
high-dimensional molecular-descriptor benchmarks), a polarization that held under a
matched greedy policy; the single exception, a low-discrimination band-gap
task with the smallest budget, only sharpens the boundary by marking where surrogate
choice ceases to matter. The practical implication for SDLs is
direct: on descriptor-based molecular and materials pools, a transfer-learning
surrogate is a more reliable default than the baseline MFGP, and on large-pool,
high-evaluation tasks it delivers this accuracy at up to roughly $45\times$ fewer FLOPs than the most expensive GP, reducing the number of expensive HF experiments needed to identify
promising candidates.

The most counter-intuitive result concerns uncertainty-driven exploration, whose value we
found to be case-dependent rather than uniformly beneficial or harmful. Probing it with a
five-member acquisition portfolio spanning the standard taxonomy (improvement-based,
optimistic, information-theoretic, and sampling-based acquisitions), we found that where the
LF signal transfers strongly and cleanly the transfer-learning surrogates perform best by
greedily exploiting their transfer-learned mean, and every uncertainty-driven acquisition
degrades, by up to an order of magnitude, when its BLR uncertainty is used
to drive exploration instead (COFs), so this catastrophic failure is a property of the
regime, not of a particular acquisition. On other high-correlation chemistry benchmarks
(FreeSolv, polarizability), by contrast, a sampling-based acquisition (Thompson sampling) beat the
greedy policy on every surrogate, while the improvement-based acquisitions won least often on average.
Calibration quality, measured by expected calibration error, did not predict
optimization performance: across all fifteen surrogates its correlation with regret was
negative on every benchmark (from $-0.88$ on COFs to $-0.08$ on HOPV15), so the
best-calibrated surrogates (the GP family) were, if anything, the worst
optimizers on the molecular-descriptor tasks. We interpret this through the strong-transfer regime
that often characterizes SDL pipelines: when an abundant LF signal already orders
candidates well, a high-quality posterior mean is on average more valuable than
exploration, and the wrong kind of exploration can be sharply counterproductive, though the
right kind (sampling-based) can still help on some benchmarks. This reframes the role of
uncertainty in MFBO surrogates: effort is better spent first on the quality of the
transferred representation, and where exploration is used, sampling-based acquisitions are
preferable to improvement-based ones, with a benefit that is benchmark-dependent.

The benchmark also clarifies when a learned surrogate is needed at all. Our
screening-regret analysis separated tasks in which a surrogate-free LF screen is
near-optimal (Regime~A) from those in which it fails despite a high global correlation
(Regime~B), and showed that regime membership cannot be inferred from the correlation
alone. Transfer-learning surrogates were the robust choice because they matched the
non-learning bound where a surrogate was unnecessary and decisively beat it where one was
essential. Transfer learning is therefore not a universally superior method but the dependable
default for the high-dimensional molecular and materials pools where SDLs operate.

The controlled fidelity-quality grid adds a methodological caution and sharpens this
claim. The apparent size of the transfer-learning advantage depends on which member
represents the GP family: against the baseline MFGP alone, transfer
learning wins everywhere, whereas against the stronger variants the final-regret gap
largely closes and what survives is faster anytime convergence and robustness
precisely where the LF source misorders the best candidates. Comparisons of
learned surrogates against ``Gaussian processes'', in our study and in the wider
literature, should therefore be read against the strength of the chosen
representative. The grid also locates our conclusions relative to guidance built on
global fidelity correlation~\citep{sabanza2025best}: because the advantage is
organized along the optimum-agreement axis and is nearly flat in $R^2$, a decision
map indexed by global correlation alone cannot separate the conditions in which a
learned surrogate is essential from those in which it is dispensable. The scope of
this evidence is deliberately narrow, a mechanism experiment with curated
LF sources on a single substrate, whose synthetic degradations need not mirror
real experimental or simulation biases; the benchmark suite, not the grid,
carries the generalization claim.

Several limitations bound these conclusions and motivate future work. Our protocol is
restricted to two fidelity levels with a deterministic round-robin schedule that reflects
the parallel-resource structure of many SDLs but does not exercise adaptive, cost-aware
fidelity selection; the residual parameterization extends naturally to hierarchical
fidelities, which we leave to future work. The transfer-learning surrogates use fixed, modest-capacity
multilayer perceptrons on precomputed descriptors reduced by PCA;
larger architectures or learned molecular representations could further widen the gap but
were deliberately excluded to keep the comparison fair. Our evaluation is also retrospective,
optimizing over fixed candidate pools that emulate the closed loop rather than driving a live
autonomous platform, so the surrogate ranking we report remains to be confirmed in
deployment. Finally, the comparison is against
the baseline MFGP and the three MFGP variants rather than the entire space
of GP methods, so our claims are scoped to this widely used family under a
pool-based, two-fidelity protocol.

Taken together, these results establish transfer-learning surrogates as the surrogate
of choice for molecular and materials multi-fidelity optimization, and yield a concrete
recipe for the engine at the core of an SDL: default to a transfer-learning surrogate on descriptor-based
molecular and materials pools, reserve GPs for the smooth, low-dimensional
design spaces where their inductive bias is an asset, and invest first in the quality of the
transferred mean, treating uncertainty-driven exploration as a case-dependent refinement
that, where used, is best supplied by a sampling-based acquisition. Because this guidance
concerns the surrogate rather than the acquisition policy or fidelity schedule, it can be
adopted within existing autonomous-experimentation platforms with no change to their
optimization policy. Extending this comparison to adaptive, cost-aware fidelity selection
and to hierarchical chains of three or more fidelities is the natural next step towards a
complete and reliable decision-making engine for autonomous discovery.

\section{Methods}\label{sec:methods}

\subsection{Problem setup}\label{sec:meth-setup}

We considered the problem of MFBO for minimizing an expensive
HF objective under a finite experimental budget.  The HF
function $f^{(H)} : \mathcal{X} \to \mathbb{R}$ represents the target objective over a compact design space $\mathcal{X} \subseteq \mathbb{R}^d$ with the aim of identifying the global minimizer $x^\star = \arg\min_{x \in \mathcal{X}} f^{(H)}(x)$. A
cheaper LF function $f^{(L)}$ provides biased but potentially informative observations of the same landscape at reduced cost. We characterized the informativeness of the LF source by its squared correlation with the HF objective, $R^2 =
\mathrm{Corr}(f^{(L)}(x), f^{(H)}(x))^2$. Evaluation cost was modelled as $c_H = 1$ for
HF and $c_L = \rho$ for LF with $\rho \ll 1$. At each step, the optimizer selected a design point 
$(x_t, s_t)$ and a fidelity level $s_t \in \{L, H\}$ subject to the total budget constraint $\sum_t c_{s_t} \le B$. The 
performance was measured using the simple regret over HF evaluations, $r_T = \min_{t:
s_t = H} f^{(H)}(x_t) - f^{(H)}(x^\star)$.

\subsection{Pool-based optimization loop and acquisition}\label{sec:meth-loop}

Optimization was performed over a fixed candidate pool $\mathcal{X}_{\mathrm{cand}}$  with sampled candidates removed after each query. The initial design used $10\%$ of the
budget and was generated by Latin-hypercube sampling for continuous domains and furthest-point
sampling for molecular pools~\citep{mckay2000comparison}. At each iteration, the surrogate model was retrained
de novo on all accumulated observations. A deterministic round-robin
schedule issued $n_{\mathrm{LF}\rightarrow\mathrm{HF}} = \lfloor 1/\rho \rfloor$
LF queries between successive HF queries, isolating surrogate quality
from cost-aware fidelity heuristics and reflecting SDL pipelines in which low- and
HF instruments operate in parallel. Candidate selection used EI~\citep{jones1998efficient} as the unified acquisition objective,
\begin{equation}\label{eq:ei}
\begin{aligned}
\alpha_{\mathrm{EI}}(x) &= (\hat{y} - \mu(x) - \xi)\,\Phi(Z) + \sigma(x)\,\phi(Z), \\
Z &= \frac{\hat{y} - \mu(x) - \xi}{\sigma(x)},
\end{aligned}
\end{equation}
where $\hat{y}$ is the incumbent value, $\xi = 0.01$~\citep{snoek2012practical}, and $\Phi$, $\phi$ denote the
standard normal CDF and PDF respectively. The acquisition function was maximized by enumeration over the pool. For the
GP baselines, query locations at both fidelity levels were selected using the cross-fidelity posterior evaluated at the
HF level, which provides a calibrated predictive standard deviation in closed
form. For the transfer-learning surrogates, the main results used the greedy policy: LF EI was computed with a constant predictive
standard deviation, so that the EI ordering reduces to greedy
selection by the BLR-regularized posterior mean, while HF candidates were selected
by the minimum predicted mean. To assess sensitivity to uncertainty modelling, we evaluated several uncertainty-driven configurations.  These included a portfolio of five LF acquisitions using the BLR predictive standard
deviation: EI, probability of
improvement~\citep{kushner1964new}, the GP lower confidence
bound~\citep{srinivas2012information}, max-value entropy search~\citep{wang2017max}, and
Thompson sampling~\citep{russo2018tutorial}, spanning the standard acquisition-function
taxonomy~\citep{shahriari2015taking}. We further tested a fully uncertainty-aware variant with a
second BLR head on the HF network and EI applied at both fidelities. These variants are analysed in
the Results (Fig.~\ref{fig:portfolio}) and reported in full in the Supplementary
Information. All objectives were formulated as minimization problems, with maximization
targets negated.

\subsection{Surrogate models}\label{sec:meth-surrogates}

Our GP baselines covered standard, nonlinear, learned-feature, and
sparse-variational regimes of multi-fidelity GP modelling. The baseline MFGP was implemented using the
\texttt{SingleTaskMultiFidelityGP} in \texttt{BoTorch}~\citep{balandat2020botorch}, corresponding to an
autoregressive Kennedy--O'Hagan formulation with a Mat\'{e}rn-5/2 kernel and fidelity
encoded as a binary indicator. Hyperparameters were fitted by exact
marginal-likelihood maximization. NARGP~\citep{perdikaris2017nonlinear} models the
HF response as a nonlinear function of the input and the LF posterior
mean. The deep-kernel GP~\citep{wilson2016deep} replaces the conventional kernel with one
defined on neural network features trained jointly with the marginal-likelihood optimization. The sparse
variational MFGP approximated the multi-fidelity posterior using stochastic variational
inference over inducing points with a multi-task index kernel, providing a scalable alternative to
exact GP inference~\citep{titsias2009variational, hensman2013gaussian}. The transfer-learning surrogates followed a two-network residual parameterization
in which an LF predictor $g_{\theta}^{(L)}:\mathbb{R}^d\to\mathbb{R}$ and an
HF residual network $g_{\phi}^{(H)}:\mathbb{R}^{d+1}\to\mathbb{R}$ (conditioned
on the input and the LF prediction) combine as
\begin{equation}\label{eq:residual}
\mu^{(H)}(x) = g_{\theta}^{(L)}(x) + g_{\phi}^{(H)}\!\left([x,\, g_{\theta}^{(L)}(x)]\right).
\end{equation}

This parameterization encourages the residual network to learn the discrepancy $\delta(x) = f^{(H)}(x) - f^{(L)}(x)$,
which is smoother than the HF function itself when the two fidelities are
correlated. Both networks were two-layer multilayer perceptrons of hidden width 64 with
$\tanh$ activations. The choice of $\tanh$ rather than ReLU was validated in a matched ablation
reported in the Supplementary Information. Predictive uncertainty was obtained from a BLR head on the final-layer features of the
LF network. This yielded a closed-form predictive mean and
variance with prior precision $\alpha = 1$ and noise precision $\beta = 1$. The 
derivation, together with the eleven transfer-learning
architectures, their training schedules, and method-specific hyperparameters is provided
in the Supplementary Information.

\subsection{Transfer-learning framework}\label{sec:meth-transfer}

All transfer-learning surrogates shared a two-stage pipeline. First, the LF
predictor was pretrained using the larger LF dataset. The model was then adapted to the
HF task using the smaller HF dataset, while keeping the backbone capacity 
fixed so that the performance differences reflected the transfer mechanism rather than model size. The mechanisms spanned five paradigms:
representation transfer including sequential fine-tuning, progressive unfreezing, and adapter
modules that update a small number of parameters over a frozen backbone; parameter transfer, including pretrain-then-joint training and soft parameter sharing; output transfer including knowledge
distillation and pseudo-labelling; data transfer implemented as curriculum learning by ordering
HF samples according to residual magnitude; distribution alignment implemented using domain adaptation by
a maximum-mean-discrepancy penalty between LF and HF feature distributions; and gradient
coupling, comparing stop-gradient and end-to-end joint training. Full loss functions and method-specific
hyperparameters are provided in the Supplementary Information.

\subsection{Benchmarks}\label{sec:meth-benchmarks}

\begin{table}[h]
\centering
\caption{Benchmark suite. $d$: input dimension, $|\mathcal{X}_{\mathrm{cand}}|$:
candidate-pool size, $\rho$: cost ratio, $R^2$: LF--HF correlation, $B$: total budget in
HF-equivalent units. HOPV15 and Matbench-gap extend the original suite of~Sabanza-Gil et
al.~\citep{sabanza2025best} with two materials tasks introduced here.}
\label{tab:benchmarks}
\begin{tabular}{lccccc}
\toprule
\textbf{Benchmark} & $d$ & $|\mathcal{X}_{\mathrm{cand}}|$ & $\rho$ & $R^2$ & $B$ \\
\midrule
Branin-Fav & 2 & 2{,}500 & 0.1 & 0.97 & 50 \\
Branin-Unfav & 2 & 2{,}500 & 0.5 & 0.56 & 50 \\
Park-Fav & 4 & 256 & 0.1 & 0.88 & 50 \\
Park-Unfav & 4 & 256 & 0.5 & 0.42 & 50 \\
COFs & 14 & 608 & 0.065 & 0.98 & 30 \\
FreeSolv & 10 & 640 & 0.1 & 0.88 & 50 \\
Polarizability & 10 & 1{,}134 & 0.167 & 0.99 & 30 \\
HOPV15 & 10 & 240 & 0.1 & 0.02 & 30 \\
Matbench-gap & 10 & 3{,}347 & 0.05 & 0.46 & 20 \\
\bottomrule
\end{tabular}
\end{table}

We evaluated all methods using nine benchmarks spanning input dimension, cost ratio, and LF--HF
correlation. The squared LF--HF correlation $R^2$ ranged from $0.02$ for organic
photovoltaics, where the LF proxy is nearly uninformative, to $0.99$ for molecular polarizability.
Key benchmark properties are summarized in Table~\ref{tab:benchmarks}. The synthetic benchmarks comprised the two-dimensional Branin function
and four-dimensional Park function, each evaluated under favourable and unfavourable fidelity relationships. In both cases, a
bias parameter $\alpha \in [0,1]$ controlled the discrepancy between LF and HF functions, with $\alpha = 1$ corresponding to identical fidelities. Full function definitions and scenario settings are given in the Supplementary
Information. The chemistry and materials benchmarks comprised five empirical datasets. COFs contained $608$ covalent organic
frameworks for Xe/Kr selectivity represented by $14$ structural descriptors with grand-canonical
Monte Carlo simulations as the HF source and Henry's-law estimates as the LF source ($\rho =
0.065$)~\citep{gantzler2023multifidelity}. FreeSolv contained $640$ molecules with experimental
and computed hydration free energies~\citep{mobley2014freesolv}. The polarizability benchmark included
$1{,}134$ molecules from the Alexandria library with experimental versus Hartree--Fock
values~\citep{ghahremanpour2018alexandria, fare2022multi}. HOPV15 contained $240$
organic-photovoltaic donor molecules, using experimental power conversion efficiency as the HF target
and a Scharber-model estimate from computed frontier orbitals as the LF proxy.
Matbench-gap contained $3{,}347$ inorganic compositions with experimental and computed band
gaps. Molecular benchmarks were encoded as RDKit or composition-based descriptors,
standardized, and reduced to ten dimensions by PCA once at
initialization; maximization targets were negated. Cost ratios were matched to the originating
studies to enable indirect comparison with prior single- and multi-fidelity GP
results~\citep{sabanza2025best}.

\subsection{Controlled fidelity-quality grid}\label{sec:meth-grid}

%% NOTE (2026-07-24 effect-size pass): verified against the grid data
%% (mlip_elastic/grid_manifest.csv + results_grid_polariz/cells): 126 cells,
%% 10 seeds each, R^2 0.10-0.91, top-10 agreement 0.00-1.00; the family
%% membership below was corrected to the models actually run on the grid
%% (TL = Progressive / Knowledge Distillation / Pseudo-Labelling; variants =
%% deep-kernel + sparse variational, no NARGP).
%% TODO(authors): still to verify: random-forest configuration, knob
%% definitions and their calibration, budget and cost ratio used for the grid
%% runs, and the binning of the displayed maps.
To vary the two properties of LF quality independently, we constructed a
grid of curated LF sources on the polarizability substrate. Each source was
derived from out-of-fold random-forest predictions of the HF objective and
perturbed by two knobs: a bulk knob that degrades the global LF--HF $R^2$, and a
gentle-demotion knob that selectively demotes top-ranked candidates, lowering the
top-10 optimum agreement (the overlap between the LF and HF top-10
candidate sets) while leaving the bulk of the ranking intact. Random-forest
predictions were chosen as the base, rather than direct perturbations of the
ground-truth values, because the errors of a learned predictor carry the smooth,
descriptor-correlated structure of a realistic LF proxy, whereas perturbing the
ground truth directly would inject noise with an arbitrary, unstructured form.
These synthetic degradations nevertheless need not reproduce the structure of
real experimental or simulation biases, which bounds how far the grid's
conclusions generalize beyond the polarizability substrate. Calibrating the two
knobs produced $126$ conditions covering $R^2 \approx 0.1$--$0.9$ and agreement
$0$--$1$, each run for ten seeds under the pool-based optimization loop of
Section~\ref{sec:meth-loop}, with the budget, cost ratio, and initial-design rule of
the polarizability benchmark (Table~\ref{tab:benchmarks}) and seeds shared across all
surrogates within a condition.

Three families were compared pairwise: the transfer-learning family, represented in
the grid by Progressive, Knowledge Distillation, and Pseudo-Labelling; the baseline
MFGP; and the MFGP variants, represented by the deep-kernel and sparse variational
models (NARGP was not run on the grid).
Family performance in a cell is that of its best member in that cell
(best-of-family), and the advantage of one family over another is the mean difference
of the metric (second-named minus first-named family; both metrics are
lower-is-better, so positive values favour the first-named family). Two metrics were
scored: final regret at the budget, and anytime performance, defined as the area
under the simple-regret-versus-budget curve, which rewards fast convergence rather
than only the endpoint. Cell-level significance was assessed with two-sided paired
Wilcoxon signed-rank tests over the ten shared seeds, reported together with the
matched-pairs rank-biserial correlation $r$ as the effect size (Methods,
``Implementation and reproducibility''; per-comparison distributions in Supplementary
Table~\ref{supptab:grid_effect_sizes}) and corrected with the Benjamini--Hochberg
(BH) false-discovery-rate procedure at $q = 0.05$ across the $126$ cells of each
comparison--metric map; comparisons with identical per-seed values enter the
correction with $p = 1$. In the displayed maps the conditions are binned by ($R^2$,
agreement); because $74$ of the $97$ non-empty bins hold a single condition and the
rest hold two or three, bin-level evidence is assessed with one test per bin rather
than a vote: a stratified signed-rank test that uses the bin's conditions as strata
and the paired per-seed differences within each condition as units (a paired-sample
analogue of the van Elteren stratified rank test), with the null distribution
obtained by sign-flip randomization (exact enumeration up to $2^{16}$ sign
assignments, otherwise $2\times10^{5}$ Monte-Carlo draws). Bin-level $p$-values are
BH-corrected across each map's non-empty bins and an asterisk marks $q \le 0.05$; in
a single-condition bin the test reduces to that condition's Wilcoxon test. The
map-wide direction of each comparison was additionally tested once with a two-sided
Wilcoxon signed-rank test across the $126$ per-cell advantages (conditions as
units), and colour scales are shared within each row of the grid. Sensitivity of the
maps to the starring criterion is reported in Supplementary
Note~\ref{suppnote:family_split}.
Because best-of-family selection can favour the family with more members, the
analysis was repeated with the unselected family mean, which preserved the ordering
of the families (Supplementary Note~\ref{suppnote:family_split}).

\subsection{Implementation and reproducibility}\label{sec:meth-impl}

All surrogates were trained using Adam and fixed $\ell_2$ weight decay, full-batch
optimization and fixed epoch budgets. Dropout, batch normalization, and gradient
clipping were not used. Models were retrained de novo at every optimization iteration to remove path dependence
from warm-starting. Inputs and targets were standardized using statistics pooled across
fidelities, and descriptor standardization and PCA were applied
once to the fixed candidate pool. Each benchmark-surrogate pair was evaluated for $20$ random seeds.
Reported values are means over seeds with uncertainty as the standard error of the mean.
Paired comparisons between surrogates use two-sided Wilcoxon signed-rank tests across
matched seeds, with zero differences discarded; where both sides reach the same regret
to the reported precision and the residual paired differences are at floating-point
scale, the comparison is recorded as a tie rather than tested. As the matched effect
size we report
the rank-biserial correlation $r = (W_{+} - W_{-})/(W_{+} + W_{-})$, where $W_{+}$ and
$W_{-}$ are the signed-rank sums of the nonzero paired differences favouring each
side. $r$ ranges from $-1$ to $1$; $r > 0$ favours the first-named side (the
transfer-learning side unless stated otherwise) and $|r| = 1$ means every decided seed
favours the same side. Where the same comparison is repeated across the benchmark
suite, the resulting $p$-values are additionally assessed under the
Benjamini--Hochberg procedure at $q = 0.05$, as they are within each map of the
controlled grid; $p$-values quoted in the text and tables are unadjusted, and the
outcome of the correction is reported in Supplementary
Table~\ref{supptab:effect_sizes}. The ordering in the average-rank panel
(Fig.~\ref{fig:final_regret}k) was assessed with the Dem\v{s}ar
procedure~\citep{demsar2006statistical}: a Friedman test with the nine benchmarks as
blocks and surrogates ranked within each benchmark by their seed-mean value (mid-ranks
for ties), followed by Nemenyi post-hoc comparisons, for which the critical difference
with fifteen models and nine benchmarks is $7.15$ rank units at $\alpha = 0.05$.
Common and
model-specific hyperparameters, GP configuration, the
BLR derivation, synthetic function definitions, and
ablation studies (including activation, uncertainty placement, acquisition variants, promotion, and
adaptive scheduling) are provided in the Supplementary Information.

%% Data availability / Code availability sections removed for the preprint:
%% "released on publication" is self-contradictory here, since the preprint is
%% itself a publication, and there is no repository URL to give yet. Reinstate
%% both sections with the Zenodo/GitHub DOI at journal submission, where they
%% are required. The three dataset citations they carried (COFs, FreeSolv,
%% Alexandria) are still made in Methods, ``Benchmarks''.

\backmatter

\bmhead{Acknowledgements}

The authors would like to acknowledge the financial support from the King's College
London Net Zero Centre Ph.D.\ Scholarship scheme.

\bibliography{references}

%%%%%%%%%%%%%%%%%%%%%%%%%%%%%%%%%%%%%%%%%%%%%%%%%%%%%%%%%%%%%%%%%%%%%%%%%%%%%%%
%% SUPPLEMENTARY INFORMATION
%% At submission this content is provided as a separate Supplementary
%% Information file. It is retained here, set single-column, as a working
%% appendix containing full reproducibility detail and the ablation studies
%% referenced from the main text.
%%%%%%%%%%%%%%%%%%%%%%%%%%%%%%%%%%%%%%%%%%%%%%%%%%%%%%%%%%%%%%%%%%%%%%%%%%%%%%%
\onecolumn

% ---------------------------------------------------------------------------
% Supplementary Information
% Replace the previous \begin{appendices}...\end{appendices} block with this block.
% ---------------------------------------------------------------------------

\section*{Supplementary Information}

% Supplementary Note counter for Nature/NCS-style references.
% This lets \ref{suppnote:...} print the Supplementary Note number.
\newcounter{suppnotecounter}
\newcommand{\suppnote}[2]{%
  \refstepcounter{suppnotecounter}%
  \section*{Supplementary Note \thesuppnotecounter. #1}%
  \label{#2}%
}

% Supplementary figure/table numbering.
% These make captions appear as "Supplementary Figure 1" and
% "Supplementary Table 1" in standard LaTeX classes.
\setcounter{figure}{0}
\setcounter{table}{0}
\renewcommand{\figurename}{Supplementary Figure}
\renewcommand{\tablename}{Supplementary Table}

\suppnote{MFGP baseline implementation details}{suppnote:mfgp}

We provide implementation details for the multi-fidelity Gaussian process (MFGP) baseline.

\subsection{Model configuration}

We use \texttt{SingleTaskMultiFidelityGP} from BoTorch~\citep{balandat2020botorch}, which
internally employs a Mat\'{e}rn 5/2 kernel with automatic relevance determination (ARD)
for the input dimensions and a linear truncated kernel for the fidelity dimension.
Supplementary Table~\ref{supptab:mfgp_config} summarizes the key configuration.

\begin{table}[h]
\centering
\caption{MFGP baseline configuration.}
\label{supptab:mfgp_config}
\begin{tabular}{ll}
\toprule
\textbf{Component} & \textbf{Specification} \\
\midrule
GP class & \texttt{SingleTaskMultiFidelityGP} \\
Input kernel & Mat\'{e}rn 5/2 + ARD \\
Fidelity kernel & Linear truncated kernel \\
Outcome transform & \texttt{Standardize(m=1)} \\
Hyperparameter optimization & Exact marginal log-likelihood via L-BFGS-B \\
Fidelity encoding & Last column: 0 = LF, 1 = HF \\
\bottomrule
\end{tabular}
\end{table}

At each iteration the GP was retrained from scratch by maximizing the exact marginal
log-likelihood with L-BFGS-B via \texttt{fit\_gpytorch\_mll}, and predictions were taken
from the posterior at the high-fidelity (HF) level $s=1$. The multi-fidelity GP uses this
single cross-fidelity posterior for acquisition at both fidelity steps.

\suppnote{Uncertainty quantification via Bayesian linear regression}{suppnote:blr}
Uncertainty for the transfer-learning surrogates is obtained from a Bayesian linear regression (BLR)
head on the low-fidelity (LF) network's last-layer features. Let $\varphi(x) \in \mathbb{R}^D$
denote the penultimate-layer representation and $\tilde{\varphi}(x) = [\varphi(x), 1]$ the
augmented feature vector, with a Gaussian prior over the weights $w$:
\begin{equation}
y \mid x \sim \mathcal{N}\!\left(\tilde{\varphi}(x)^\top w,\, \beta^{-1}\right), \qquad w \sim \mathcal{N}\!\left(0,\, \alpha^{-1} I\right),
\end{equation}
with prior precision $\alpha$ and noise precision $\beta$. The closed-form posterior
yields predictive moments
\begin{equation}
\mu(x) = \tilde{\varphi}(x)^\top m_N, \qquad \sigma^2(x) = \beta^{-1} + \tilde{\varphi}(x)^\top S_N\, \tilde{\varphi}(x),
\end{equation}
where $(m_N, S_N)$ are the posterior mean and covariance and we set $\alpha = \beta = 1$.
In the primary configuration the LF expected improvement (EI) uses a constant predictive standard
deviation, so the BLR head contributes a regularized posterior mean and the acquisition
reduces to greedy selection; the five-acquisition uncertainty-driven portfolio that instead
uses the BLR predictive standard deviation, and an optional HF BLR head with EI
on both fidelities, are evaluated in Supplementary Note~\ref{suppnote:acquisition_variants}.

\suppnote{Implementation details}{suppnote:implementation}

All transfer-learning surrogates share a common architectural backbone, with model-specific training
configurations detailed below.

\subsection{Common settings}
\label{suppsec:common_settings}
Supplementary Table~\ref{supptab:common_hyperparams} summarizes the shared hyperparameters across all deep
surrogates.

\begin{table}[h]
\centering
\caption{Shared hyperparameters across all transfer-learning surrogates.}
\label{supptab:common_hyperparams}
\begin{tabular}{ll}
\toprule
\textbf{Parameter} & \textbf{Value} \\
\midrule
Hidden dimension & 64 \\
Number of layers & 2 (for each of LF and HF networks) \\
Activation & Tanh \\
Optimizer & Adam \\
$\ell_2$ weight decay & 1e-4 \\
BLR prior precision $\alpha$ & 1.0 \\
BLR noise precision $\beta$ & 1.0 \\
\bottomrule
\end{tabular}
\end{table}

\subsection{Model-specific settings}
\label{suppsec:model_settings}
Supplementary Table~\ref{supptab:model_hyperparams} presents the training hyperparameters for each transfer-learning
method. Sequential methods use 200 LF epochs followed by 100 HF epochs, while
joint training methods use 300 epochs for simultaneous LF and HF optimization.

\begin{table}[h]
\centering
\caption{Training hyperparameters for each transfer-learning method. LR denotes learning
rate.}
\label{supptab:model_hyperparams}
{\footnotesize%
\begin{tabular}{lccccl}
\toprule
\textbf{Model} & \textbf{LF LR} & \textbf{HF LR} & \textbf{LF Epochs} & \textbf{HF Epochs} & \textbf{Method-Specific Parameters} \\
\midrule
Sequential & 1e-3 & 1e-3 & 200 & 100 & LF frozen during HF training \\
Progressive & 1e-3 & 1e-3 $\rightarrow$ 1e-4 & 200 & 50+50 & Stage 1: head only; Stage 2: full finetune \\
Curriculum & 1e-3 & 1e-3 & 200 & 100 & HF samples added gradually \\
Pretrain-then-Joint & 1e-3 & 1e-4 & 200 & 100 & Stage 1: LF only; Stage 2: joint training \\
Stop-Gradient Joint & 1e-3 & 1e-3 & 300 & (joint) & $\alpha_{\text{loss}}$=0.5; stop-gradient on LF$\rightarrow$HF \\
End-to-End Joint & 1e-3 & 1e-3 & 300 & (joint) & Full gradient flow \\
Soft Parameter Sharing & 1e-3 & 1e-3 & 300 & (joint) & $\lambda_{\text{soft}}$=0.01 \\
Knowledge Distillation & 1e-3 & 1e-3 & 200 & 100 & $\alpha_{\text{KD}}$=0.3; $T$=3.0 \\
Domain Adaptation (MMD) & 1e-3 & 1e-3 & 200 & 100 & $\lambda_{\text{MMD}}$=0.1 \\
Pseudo-Labelling & 1e-3 & 1e-3 & 200 & 100 & $w_{\text{pseudo}}$=0.5 \\
Adapter & 1e-3 & 1e-3 & 200 & 100 & $d_{\text{bottleneck}}$=16; backbone frozen \\
\bottomrule
\end{tabular}%
}
\end{table}

\subsection{Training procedures and method-specific parameters}
\label{suppsec:training_procedures}
All models use $\ell_2$ weight decay (WD=1e-4). \emph{Sequential}: the LF network is
trained first and frozen, then the HF residual network is trained on top.
\emph{Progressive}: Stage 1 trains only the HF head with the backbone frozen; Stage 2
finetunes the full network at a reduced learning rate. \emph{Curriculum}: at epoch $e$ the
number of HF samples used is $n_{\text{use}} = \min(n, \max(2, e/200 \cdot n))$.
\emph{Pretrain-then-Joint}: Stage 1 pretrains on LF data; Stage 2 jointly trains on both
fidelities at a reduced learning rate. \emph{Stop-Gradient Joint}: joint training with
stop-gradient on the LF network and loss weighting $\mathcal{L} = \alpha_{\text{loss}}
\mathcal{L}_{\text{LF}} + (1-\alpha_{\text{loss}})\mathcal{L}_{\text{HF}}$,
$\alpha_{\text{loss}}=0.5$. \emph{End-to-End Joint}: joint training with full gradient
flow. \emph{Soft Parameter Sharing}: penalty
$\lambda_{\text{soft}}\|\theta^{(L)}-\theta^{(H)}\|_2^2$, $\lambda_{\text{soft}}=0.01$.
\emph{Knowledge Distillation}: HF student trained with task loss plus distillation loss,
$\alpha_{\text{KD}}=0.3$, $T=3.0$. \emph{Domain Adaptation (MMD)}: maximum-mean-discrepancy penalty
$\lambda_{\text{MMD}}=0.1$ aligning LF and HF features. \emph{Pseudo-Labelling}: LF
pseudo-labels weighted $w_{\text{pseudo}}=0.5$. \emph{Adapter}: frozen LF backbone with a
bottleneck adapter ($d_{\text{bottleneck}}=16$) and HF head.

\suppnote{Synthetic function definitions}{suppnote:synthetic_functions}

Both synthetic functions are standard multi-fidelity Bayesian optimization (MFBO) benchmarks adopted from~Sabanza-Gil et
al.~\citep{sabanza2025best}.

\subsection{Branin function}
The HF Branin function is
\begin{equation}
f^{(H)}(x_1, x_2) = \left(x_2 - b x_1^2 + c x_1 - 6\right)^2 + 10\left(1 - \frac{1}{8\pi}\right)\cos(x_1) + 10,
\end{equation}
with $b = 5.1/(4\pi^2)$, $c = 5/\pi$, domain $x_1 \in [-5, 10]$, $x_2 \in [0, 15]$, and
global minimum $f^* = 0.397887$. The LF approximation modifies only the
quadratic coefficient, $b_{\text{LF}} = b - 0.1(1-\alpha)$, so that the correlation
degrades as $\alpha$ decreases.

\subsection{Park function}
The HF Park function is
\begin{equation}
f^{(H)}(x_1, x_2, x_3, x_4) = \frac{x_1}{2}\sqrt{1 + \frac{(x_2 + x_3^2)x_4}{x_1^2}} + (x_1 + 3x_4)\exp\!\left(1 + \sin(x_3)\right),
\end{equation}
with $x_i \in [0,1]$ and global minimum $f^* = 0$. The LF approximation modifies
the $x_4$ coefficient ($3 \rightarrow 3 - 1.5(1-\alpha)$) and scales the exponential
argument by $\alpha$.

\subsection{Scenario configurations}
Supplementary Table~\ref{supptab:synthetic_scenarios} lists the $\alpha$ values and resulting
characteristics. The $R^2$ values were computed by sampling 1,000 uniform points,
evaluating both fidelities, and fitting a linear regression from LF to HF. Candidate pools
were constructed by uniform grid discretization ($50^2 = 2{,}500$ for Branin; $4^4 = 256$
for Park).

\begin{table}[h]
\centering
\caption{Synthetic function scenario configurations.}
\label{supptab:synthetic_scenarios}
\begin{tabular}{llccccc}
\toprule
\textbf{Benchmark} & \textbf{Scenario} & $\alpha$ & $R^2$ & $\rho$ & \textbf{Pool Size} & $f^*$ \\
\midrule
Branin-Fav & Favourable & 0.8 & 0.97 & 0.1 & 2,500 & 0.398 \\
Branin-Unfav & Unfavourable & 0.1 & 0.56 & 0.5 & 2,500 & 0.398 \\
Park-Fav & Favourable & 0.6 & 0.88 & 0.1 & 256 & 0.0 \\
Park-Unfav & Unfavourable & 0.0 & 0.42 & 0.5 & 256 & 0.0 \\
\bottomrule
\end{tabular}
\end{table}

\suppnote{Ablation studies}{suppnote:acquisition_variants}

This section specifies the protocol for each ablation summarized in the main text; the
complete per-benchmark results are provided as Supplementary Data.

\paragraph{Acquisition variants.} Beyond the primary greedy policy, we evaluated a
portfolio of five uncertainty-driven acquisitions spanning the standard taxonomy of myopic
acquisition functions~\citep{shahriari2015taking}: the improvement-based EI~\citep{jones1998efficient} and probability of
improvement~\citep{kushner1964new}; the optimistic Gaussian-process lower confidence
bound~\citep{srinivas2012information} ($\beta=2$); the information-theoretic max-value
entropy search~\citep{wang2017max} (Gumbel approximation, ten max-value samples); and
sampling-based Thompson sampling~\citep{russo2018tutorial}. Each replaces only the
LF acquisition, using the BLR predictive standard deviation with the surrogate,
round-robin schedule, HF step, and seeds held fixed, and was run across all
eleven transfer-learning surrogates and nine benchmarks ($11\times 9\times 20$ seeds per
acquisition). For each acquisition and benchmark we report the fraction of surrogates on
which it beats the greedy policy: the greedy policy beat all five acquisitions on COFs on ten of the
eleven surrogates, whereas Thompson sampling beat the greedy policy on every surrogate on FreeSolv and
polarizability, and the improvement-based acquisitions ranked weakest overall
(Fig.~\ref{fig:portfolio}). We additionally evaluated a fully uncertainty-aware variant
that adds a second BLR head on the HF network and applies EI at both fidelities.
As a matched control for the surrogate-versus-acquisition question, the baseline MFGP was also run
with a greedy policy that selects by its posterior mean, all other settings held fixed
(Fig.~\ref{fig:acquisition}).

\paragraph{Activation.} To isolate the effect of the activation function, the benchmark
suite was repeated with ReLU in place of $\tanh$ under otherwise identical architectures
and training, recording expected calibration error, negative log-likelihood, sharpness,
and final regret for both.

\paragraph{Calibration.} For every surrogate we computed the LF expected
calibration error, negative log-likelihood, and sharpness on held-out candidates and
related them to the final optimization regret; the resulting correlation is reported in
the main text (Fig.~\ref{fig:calibration}).

\paragraph{Uncertainty placement, promotion, and adaptive scheduling.} Three other
ablations are detailed in the Supplementary Information: the placement of the BLR head
(LF only, HF only, or both), allowing LF candidates to be
promoted to HF rather than excluded once sampled, and replacing the round-robin
schedule with an adaptive cost-weighted policy. Each was evaluated on a representative
subset of three to five benchmarks spanning the favourable and unfavourable regimes, and on
every benchmark tested none altered the qualitative ranking of the surrogate families.

\suppnote{Computational cost: absolute FLOPs and compute ranking}{suppnote:computational_cost}

Figure~\ref{fig:scaling_law} establishes how surrogate compute scales with problem size;
here we report the absolute cost that underlies those laws. The total floating-point operations (FLOPs) of the
MFBO loop (one surrogate fit plus one pool-wide prediction per iteration,
summed over the loop) are shown per surrogate and benchmark, confirming that the
$\mathcal{O}(N^3)$ Gaussian processes dominate the budget while the $\mathcal{O}(N)$
transfer-learning surrogates stay one to two orders of magnitude cheaper.

\begin{figure*}[t]
\centering
\includegraphics[width=\textwidth]{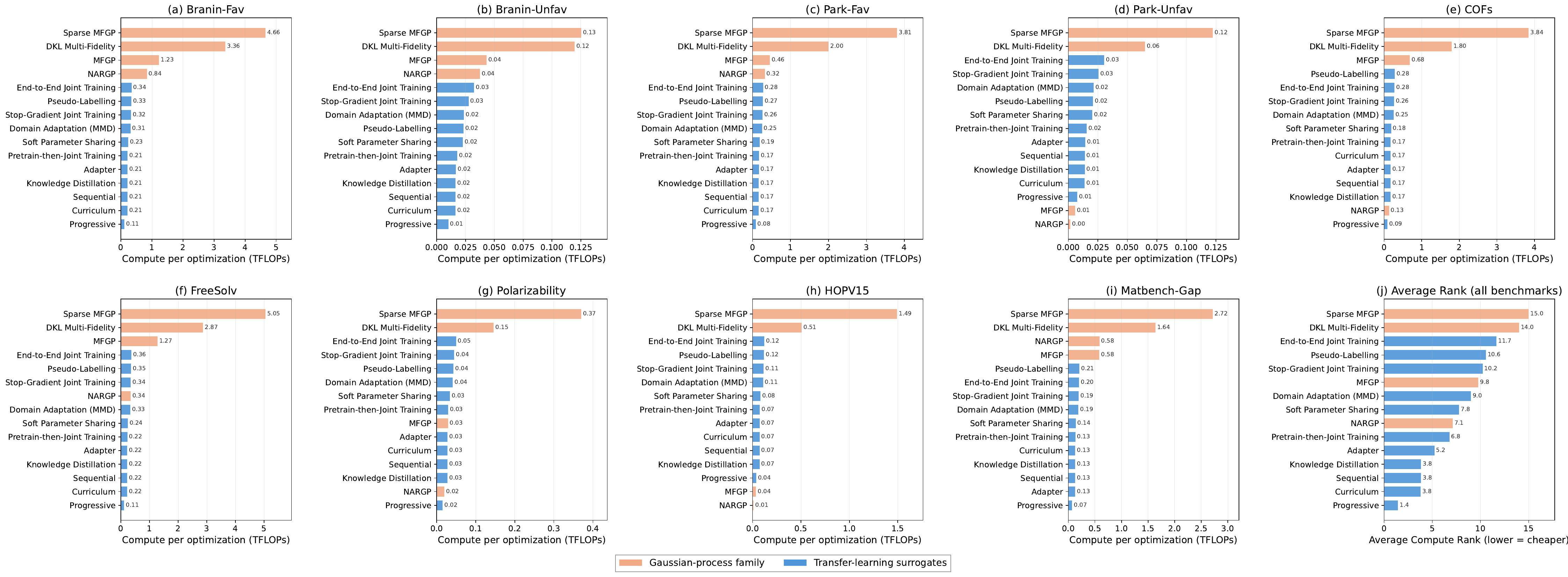}
\caption{Absolute computational cost across the nine benchmarks, in floating-point
operations (FLOPs), a hardware-independent measure: total FLOPs of the MFBO
loop (one surrogate fit plus one pool-wide prediction per iteration, summed over the loop),
with the Gaussian-process family in salmon and the transfer-learning surrogates in blue.
The sparse variational (Sparse MFGP) and deep-kernel (DKL) GPs are the most expensive on
every benchmark, up to roughly $45\times$ and $25\times$ the fastest transfer-learning surrogate; the
baseline MFGP and NARGP scale as $\mathcal{O}(N^3)$ in the number of observations (measured
fit-FLOP exponent $k = 3.0$) against $\mathcal{O}(N)$ for the transfer-learning surrogates ($k = 1.0$),
so the gap widens with the evaluation budget. Panel \textbf{j} summarizes these costs as
the average compute rank across all nine benchmarks (lower is cheaper), with the colour
split repeated in a shared legend below the panels.}
\label{suppfig:computing_time}
\end{figure*}

\suppnote{Family-split advantage maps: complete grid and robustness}{suppnote:family_split}

\begin{figure*}[t]
\centering
\includegraphics[width=0.9\textwidth]{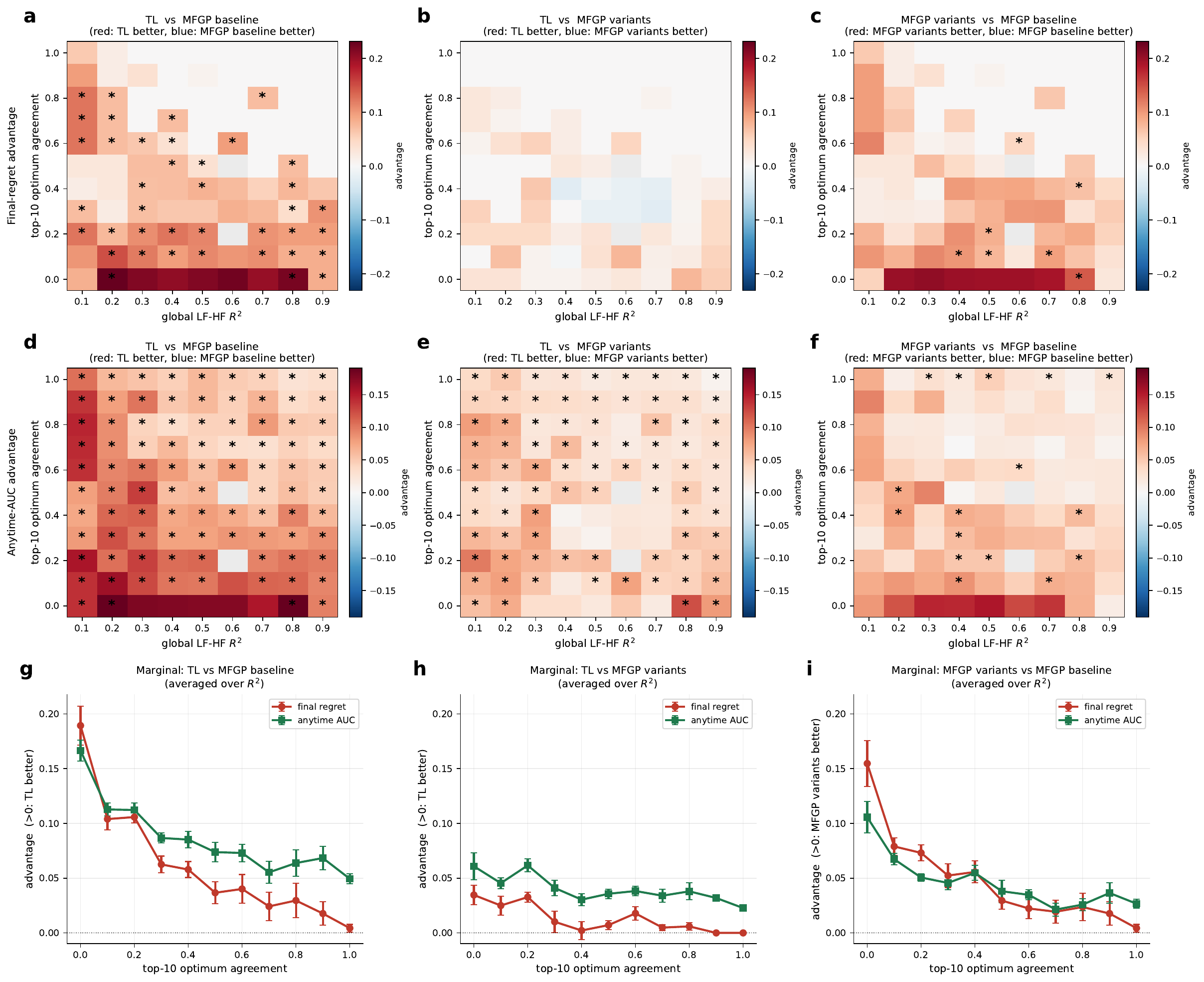}
\caption{Family-split advantage maps: the complete grid. Columns compare transfer
learning (TL) against the baseline MFGP (left), TL against the MFGP variants
(deep-kernel and sparse variational; centre), and the MFGP variants against the
baseline MFGP (right). Rows show the final-regret advantage (\textbf{a}--\textbf{c}),
the anytime-AUC advantage (\textbf{d}--\textbf{f}; the row reproduced as
Fig.~\ref{fig:family_heatmap}l--n), and the marginal advantage profiles averaged over
$R^2$ (\textbf{g}--\textbf{i}; mean $\pm$ s.e.). Within each
map, bins are indexed by the global LF--HF $R^2$ (horizontal) and the top-10 optimum
agreement between fidelities (vertical); red favours the first-named family and
asterisks mark bins significant under the per-bin stratified signed-rank test with
Benjamini--Hochberg false-discovery-rate correction across each map (Methods).}
\label{suppfig:family_split}
\end{figure*}

Supplementary Fig.~\ref{suppfig:family_split} reports the complete family-split
advantage grid underlying panels l--n of Fig.~\ref{fig:family_heatmap}
(protocol in Methods, Section ``Controlled fidelity-quality grid''): the
final-regret advantage maps, the anytime-AUC (area under the regret--budget curve) advantage maps (the row reproduced in
the main text), and the marginal advantage profiles obtained by averaging over the
global LF--HF $R^2$. The two metrics agree on the structure, with the advantage
largest at low top-10 optimum agreement and nearly flat across the $R^2$ range; the
final-regret maps additionally show that the MFGP variants close most of the baseline
MFGP's endpoint gap, whereas the anytime gap persists.

\begin{figure*}[t]
\centering
\includegraphics[width=\textwidth]{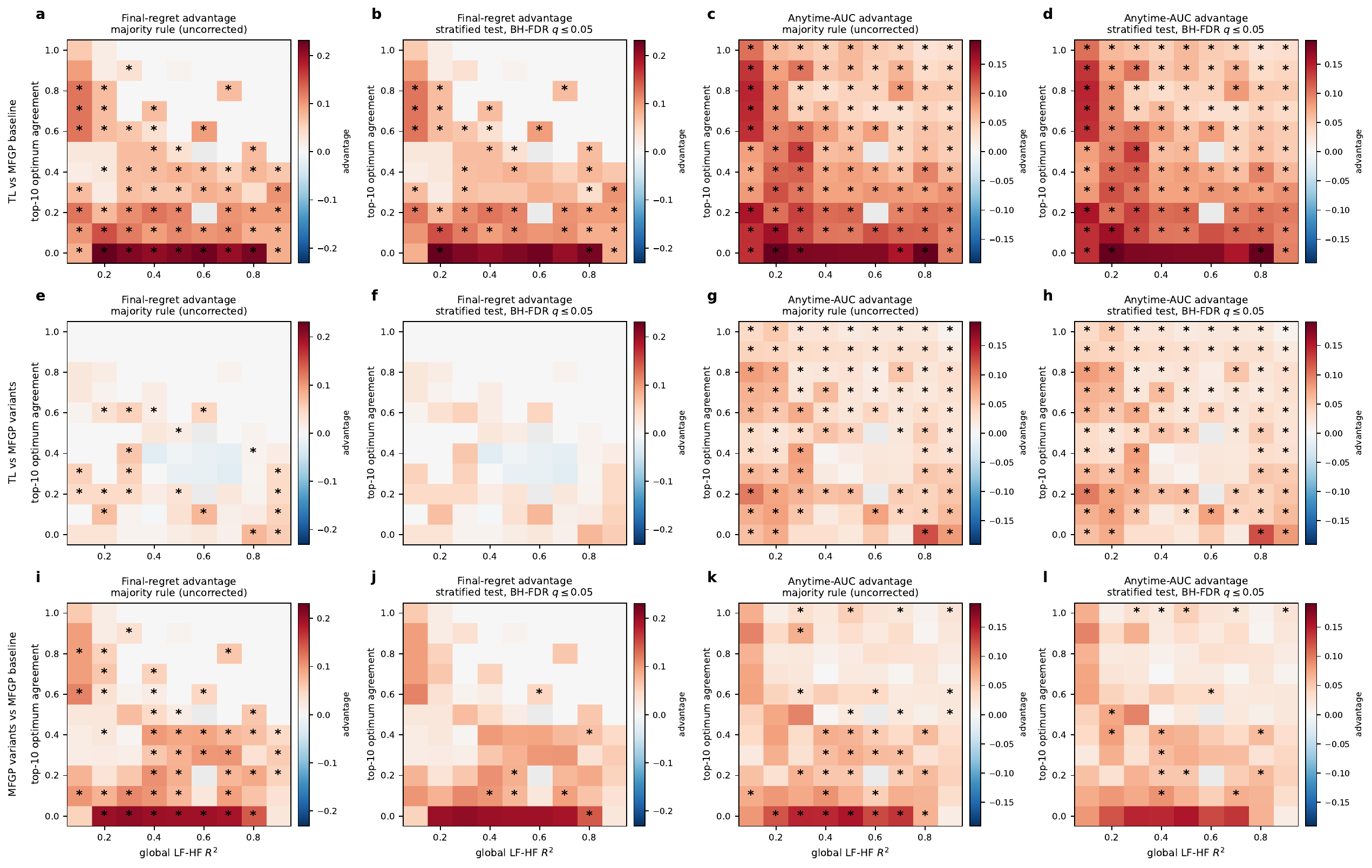}
\caption{Sensitivity of the family-split maps to the bin-starring criterion. Each row
is one pairwise comparison (top to bottom: TL versus the baseline MFGP, TL versus the
MFGP variants, and the MFGP variants versus the baseline MFGP); columns show the
final-regret advantage (\textbf{a},\textbf{b}, \textbf{e},\textbf{f},
\textbf{i},\textbf{j}) and the anytime-AUC advantage (\textbf{c},\textbf{d},
\textbf{g},\textbf{h}, \textbf{k},\textbf{l}), each under the previous uncorrected
majority rule (left member of each pair) and the corrected criterion (right member;
one stratified signed-rank test per non-empty bin, Benjamini--Hochberg
false-discovery-rate correction at $q \le 0.05$ across each map's bins; Methods).
Colours (binned mean advantage, red favouring the first-named family) are identical
within each pair; only the asterisks change. Summary counts are given in
Supplementary Table~\ref{supptab:star_sensitivity}.}
\label{suppfig:star_sensitivity}
\end{figure*}

Because best-of-family scoring can favour the family with more members (three
transfer-learning surrogates against one baseline and two variants), the analysis was
repeated with the unselected family mean; this preserved the ordering of the families
and the structure along the agreement axis.
The starring criterion itself was audited for sensitivity. The maps were originally
starred by an uncorrected majority rule (a bin starred when at least half of its
conditions reached unadjusted $p<0.05$), which ignores multiplicity across the map
and, in the $74$ single-condition bins, reduces to a single uncorrected test.
Supplementary Fig.~\ref{suppfig:star_sensitivity} and Supplementary
Table~\ref{supptab:star_sensitivity} compare this rule with the corrected criterion
adopted throughout (one stratified signed-rank test per non-empty bin, with the
bin's conditions as strata, followed by Benjamini--Hochberg false-discovery-rate
correction at $q = 0.05$ across each map's bins; Methods). The correction leaves the
transfer-learning anytime-AUC maps nearly unchanged ($94 \to 91$ and $82 \to 82$
bins starred against the baseline MFGP and the MFGP variants respectively) while
thinning the weaker maps (MFGP variants versus baseline anytime-AUC $34 \to 16$;
final regret $56 \to 39$, $20 \to 0$, and $43 \to 7$ across the three comparisons),
and the cell-level fractions quoted in the main text move from $97\%$ to $95\%$ and
from $83\%$ to $80\%$. The direction of every map is unaffected: the map-level
Wilcoxon across the $126$ per-cell advantages gives $p \le 1.7\times10^{-7}$ for all
six maps ($p = 2.0\times10^{-22}$ for each anytime-AUC comparison). The contrast
between the vanishing per-cell counts on final regret and these small map-level values
is a matter of statistical power rather than of conflicting evidence: ten seeds per
cell cannot clear a $126$-way correction unless the within-cell separation is close to
total, as it is on the anytime metric but not at the endpoint, whereas the map-level
test draws instead on the consistency of the advantage across the $126$ conditions.
Per-cell significance is therefore quoted only for the anytime maps, which carry the
claim; the final-regret maps are read from their direction and effect sizes.

%% TODO(authors): insert here, from the grid-analysis outputs:
%%   (i)   the family-mean (no-selection) maps and their full numbers;
%%   (ii)  grid-curation detail: knob-to-(R^2, agreement) calibration curves,
%%         the cell manifest, and the coverage scatter of the curated conditions;
%%   (iii) the per-cell significance table, including which member was
%%         best-of-family in each cell (per-cell and per-bin CSVs:
%%         experiments/regime_review_20260625/grid_fdr_percell.csv / _perbin.csv).

\begin{table}[h]
\centering
\caption{Sensitivity of the controlled-grid maps to the bin-starring criterion. For
each comparison--metric map of the family-split grid ($126$ cells, ten seeds per
cell): cells with a positive advantage of the first-named family; cells individually
significant under the uncorrected per-cell Wilcoxon test ($p<0.05$) and under
Benjamini--Hochberg (BH) false-discovery-rate correction across the map's $126$
cells ($q\le0.05$); display bins starred under the previous majority rule versus the
corrected criterion (one stratified signed-rank test per bin, BH across the map's
$97$ non-empty bins); and the map-level two-sided Wilcoxon $p$ across the $126$
per-cell advantages (conditions as units). The map-level $p$ saturates once every
non-tied cell favours the same family, so maps that share a sign pattern share a
$p$-value; these values express the consistency of the direction, not the size of the
advantage.}
\label{supptab:star_sensitivity}%
\begin{tabular}{@{}llccccc@{}}
\toprule
Metric & Comparison & Positive & Raw $p<0.05$ & BH $q\le0.05$ & Bins starred & Map-level $p$ \\
\midrule
Final regret & TL vs baseline MFGP & 81/126 & 65 (52\%) & 0 (0\%) & 56 $\to$ 39 & $5.3\times10^{-15}$ \\
Final regret & TL vs MFGP variants & 51/126 & 22 (17\%) & 0 (0\%) & 20 $\to$ 0 & $1.7\times10^{-7}$ \\
Final regret & MFGP variants vs baseline MFGP & 81/126 & 50 (40\%) & 0 (0\%) & 43 $\to$ 7 & $5.3\times10^{-15}$ \\
\midrule
Anytime AUC & TL vs baseline MFGP & 126/126 & 122 (97\%) & 120 (95\%) & 94 $\to$ 91 & $2.0\times10^{-22}$ \\
Anytime AUC & TL vs MFGP variants & 126/126 & 105 (83\%) & 101 (80\%) & 82 $\to$ 82 & $2.0\times10^{-22}$ \\
Anytime AUC & MFGP variants vs baseline MFGP & 126/126 & 39 (31\%) & 0 (0\%) & 34 $\to$ 16 & $2.0\times10^{-22}$ \\
\bottomrule
\end{tabular}
\end{table}

\suppnote{Omnibus rank analysis of the average-rank panel}{suppnote:rank_stats}

Panel k of Fig.~\ref{fig:final_regret} summarizes the fifteen surrogates as an average
rank across the nine benchmarks. We assessed whether that ordering is statistically
resolvable using the Dem\v{s}ar procedure (Methods, ``Implementation and
reproducibility''): a Friedman omnibus test with the benchmarks as blocks and the
surrogates ranked within each benchmark, followed, if the omnibus rejects, by Nemenyi
post-hoc comparisons.

The omnibus does not reject rank homogeneity ($\chi^2_{14} = 20.9$, $p = 0.11$), so no
pairwise ordering is claimed. The Nemenyi critical difference,
$\mathrm{CD} = q_{0.05}\sqrt{k(k+1)/(6N)}$ with $k = 15$ surrogates and $N = 9$
benchmarks, is $7.15$ rank units, which is larger than the entire observed range of
average ranks ($5.28$ for Curriculum to $11.8$; range $6.52$). The spread of the panel
is therefore smaller than the smallest difference the post-hoc test can resolve, so no
pair of surrogates is separable at this suite size, including the best against the
worst. The dashed line in Fig.~\ref{fig:final_regret}k marks this critical difference.

This null is what the benchmark design predicts rather than evidence against a
surrogate effect. The suite deliberately spans regimes in which the ranking reverses:
the Gaussian-process family is best on the two Branin functions and worst on the four
molecular-descriptor benchmarks (Fig.~\ref{fig:final_regret}b--i). An omnibus test for
a single consistent ordering across all nine benchmarks therefore tests a hypothesis
that our central result already denies, and averaging ranks over a heterogeneous suite
cancels an effect whose sign depends on the regime. Restricting the analysis to the
five chemistry and materials benchmarks does not recover resolution, because the
critical difference grows as the number of blocks falls ($\mathrm{CD} = 9.59$ at
$N = 5$).

The statistical support for the surrogate comparison accordingly rests on the
per-benchmark paired tests and effect sizes of Supplementary
Table~\ref{supptab:effect_sizes}, and on the controlled fidelity-quality grid
(Supplementary Table~\ref{supptab:grid_effect_sizes}), rather than on the pooled
ranking; Fig.~\ref{fig:final_regret}k is presented as a descriptive summary. The
omnibus null therefore indicates that no single ordering holds across a deliberately
heterogeneous suite, not that the choice of surrogate is without effect: on the
individual molecular-descriptor benchmarks the paired comparisons are significant with
near-unanimous effect sizes ($r = 0.97$--$1.00$; Supplementary
Table~\ref{supptab:effect_sizes}).

\suppnote{Effect sizes for the paired surrogate comparisons}{suppnote:effect_sizes}

Supplementary Table~\ref{supptab:effect_sizes} reports, for each of the nine
benchmarks, the paired comparison between the best transfer-learning surrogate
(chosen by mean final regret) and the baseline MFGP over the twenty matched seeds:
the matched-pairs rank-biserial correlation $r$ (Methods), the seed-level
wins/ties/losses, and the two-sided Wilcoxon $p$-value. Supplementary
Table~\ref{supptab:grid_effect_sizes} summarizes the distribution of the per-cell
$r$ over the $126$ cells of the controlled fidelity-quality grid for the three
pairwise family comparisons, on both metrics. On the anytime metric the typical
cell is unanimous: the median per-cell $r$ is $1.00$ against both GP families. On
final regret a large fraction of cells tie exactly (both sides reach regret $0$);
these ties are reported separately.

\begin{table}[h]
\centering
\caption{Per-benchmark effect sizes: best transfer-learning surrogate against the
baseline MFGP on final regret ($20$ paired seeds). The best transfer-learning (TL)
surrogate is chosen by mean final regret; $r$ is the matched-pairs rank-biserial
correlation ($r > 0$ favours transfer learning), w/t/l counts transfer-learning
wins, ties, and losses over the paired seeds, and $p$ is the two-sided Wilcoxon
$p$-value on the nonzero paired differences. Park-Fav ties on all $20$ seeds (both
sides at regret $\approx 0$); on Park-Unfav ($\dag$) both sides likewise reach regret
$\approx 0$ and the paired differences are at floating-point scale ($\sim10^{-8}$), so
the nominal comparison ($r = +0.96$, 17/2/1, $p = 2.9\times10^{-4}$) resolves numerical
noise and is not reported as an effect. Across the seven benchmarks that yield a test,
Benjamini--Hochberg correction at $q = 0.05$ leaves every comparison significant except
Matbench-gap, which is not significant before correction either.}
\label{supptab:effect_sizes}
\begin{tabular}{llrrrcr}
\toprule
Benchmark & Best TL surrogate & TL & MFGP & $r$ & w/t/l & $p$ \\
\midrule
Branin-Fav & Curriculum & 0.42 & 0.07 & $-0.79$ & 2/0/18 & $1.0\times10^{-3}$ \\
Branin-Unfav & Curriculum & 1.24 & 0.05 & $-1.00$ & 0/1/19 & $1.3\times10^{-4}$ \\
Park-Fav & Sequential & 0.00 & 0.00 & tie & 0/20/0 & -- \\
Park-Unfav & Pseudo-Labelling & 0.00 & 0.00 & tie$^{\dag}$ & -- & -- \\
COFs & End-to-End Joint Training & 0.41 & 6.55 & $+0.97$ & 18/1/1 & $2.1\times10^{-4}$ \\
FreeSolv & End-to-End Joint Training & 0.28 & 0.80 & $+1.00$ & 20/0/0 & $1.9\times10^{-6}$ \\
Polarizability & Sequential & 0.06 & 0.17 & $+1.00$ & 15/5/0 & $6.3\times10^{-4}$ \\
HOPV15 & Knowledge Distillation & 1.89 & 3.09 & $+0.61$ & 13/1/6 & $0.020$ \\
Matbench-Gap & Progressive & 0.23 & 0.29 & $+0.19$ & 9/0/11 & $0.47$ \\
\bottomrule
\end{tabular}
\end{table}

\begin{table}[h]
\centering
\caption{Effect-size distributions on the controlled fidelity-quality grid ($126$
cells, ten seeds per cell). For each pairwise family comparison and metric: the
median and interquartile range (IQR) of the per-cell matched-pairs rank-biserial
$r$ ($r > 0$ favours the first-named family), and the number of all-tie cells (every
seed pair identical, typically both sides at regret $0$; excluded from the median
and IQR). Within a cell, $r$ measures how consistently the ten paired seeds favour
one family, not how large the advantage is, so a median $r$ of $+1.00$ is compatible
with a mean advantage near zero. Cell counts, significance before and after
Benjamini--Hochberg correction, and the map-level tests are given in Supplementary
Table~\ref{supptab:star_sensitivity}.}
\label{supptab:grid_effect_sizes}
\begin{tabular}{llrrr}
\toprule
Metric & Comparison & Median $r$ & IQR & Tie cells \\
\midrule
Final regret & TL vs baseline MFGP & $+1.00$ & $[+1.00, +1.00]$ & 45 \\
Final regret & TL vs MFGP variants & $+1.00$ & $[+0.33, +1.00]$ & 60 \\
Final regret & MFGP variants vs baseline MFGP & $+1.00$ & $[+1.00, +1.00]$ & 45 \\
Anytime AUC & TL vs baseline MFGP & $+1.00$ & $[+1.00, +1.00]$ & 0 \\
Anytime AUC & TL vs MFGP variants & $+1.00$ & $[+0.91, +1.00]$ & 0 \\
Anytime AUC & MFGP variants vs baseline MFGP & $+0.67$ & $[+0.50, +0.83]$ & 0 \\
\bottomrule
\end{tabular}
\end{table}

\end{document}